\documentclass[]{fairmeta}
\usepackage{xspace}
\definecolor{cvprblue}{rgb}{0.21,0.49,0.74}

\makeatletter
\DeclareRobustCommand\onedot{\futurelet\@let@token\@onedot}
\def\@onedot{\ifx\@let@token.\else.\null\fi\xspace}

\def\eg{\emph{e.g}\onedot} 
\def\ie{\emph{i.e}\onedot}

\makeatother

\usepackage{multirow}
\usepackage{amsfonts}
\usepackage[table,xcdraw]{xcolor}
\usepackage[normalem]{ulem}
\useunder{\uline}{\ul}{}

\newcommand{\tocite}[1]{\textcolor{red}{[TO CITE]}}

\newcommand{\model}{\textsc{Tuna}\xspace}
\title{\textcolor{metablue}{Tuna}: \textcolor{metablue}{T}aming \textcolor{metablue}{U}nified Visual Representations for \textcolor{metablue}{Na}tive Unified Multimodal Models}

\author[1,2,\dagger]{Zhiheng Liu}
\author[1,3,\dagger]{Weiming Ren}
\author[1,4,\ddagger]{Haozhe Liu}
\author[1,\ddagger]{Zijian Zhou}
\author[1,\ddagger]{Shoufa Chen}
\author[1]{Haonan Qiu}
\author[1]{Xiaoke Huang}
\author[1]{Zhaochong An}
\author[1]{Fanny Yang}
\author[1]{Aditya Patel}
\author[1]{Viktar Atliha}
\author[1]{Tony Ng}
\author[1]{Xiao Han}
\author[1]{Chuyan Zhu}
\author[1]{Chenyang Zhang}
\author[1]{Ding Liu}
\author[1]{Juan-Manuel Perez-Rua}
\author[1]{Sen He}
\author[4]{Jürgen Schmidhuber}
\author[3]{Wenhu Chen}
\author[2]{Ping Luo}
\author[1]{Wei Liu}
\author[1]{Tao Xiang}
\author[1,*]{Jonas Schult}
\author[1,*]{Yuren Cong}

\affiliation[1]{Meta BizAI}
\affiliation[2]{HKU}
\affiliation[3]{University of Waterloo}
\affiliation[4]{KAUST}
\contribution[\dagger]{Joint first authors, listed alphabetically by last name}
\contribution[\ddagger]{Core contributors}
\contribution[*]{Joint project lead}

\abstract{Unified multimodal models (UMMs) aim to jointly perform multimodal understanding and generation within a single framework. We present \model, a native UMM that builds a unified continuous visual representation by cascading a VAE encoder with a representation encoder. This unified representation space allows end-to-end processing of images and videos for both understanding and generation tasks. Compared to prior UMMs with decoupled representations, \model's unified visual space avoids representation format mismatches introduced by separate encoders, outperforming decoupled alternatives in both understanding and generation. Moreover, we observe that stronger pretrained representation encoders consistently yield better performance across all multimodal tasks, highlighting the importance of the representation encoder. Finally, in this unified setting, jointly training on both understanding and generation data allows the two tasks to benefit from each other rather than interfere. Our extensive experiments on multimodal understanding and generation benchmarks show that \model achieves state-of-the-art results in image and video understanding, image and video generation, and image editing, demonstrating the effectiveness and scalability of its unified representation design.

}

\correspondence{\email{zhihengl0528@connect.hku.hk}, \email{w2ren@uwaterloo.ca},
\email{schult@meta.com},
\email{yuren@meta.com}
}

\metadata[Project page]{\url{https://tuna-ai.org}}
\vspace{8mm}

\begin{document}

\maketitle

\section{Introduction}
\label{sec:intro}
A long-term aspiration of multimodal AI is \textit{natively}\footnote{We define \textit{native} UMMs as models that are pretrained jointly on both understanding and generation objectives, instead of assembling understanding-only and generation-only models with learnable connectors. We refer to the latter as \textit{composite} UMMs.} unified multimodal generation, in which a single model can seamlessly understand and generate diverse modalities such as text, images, and videos. Recent advances in unified multimodal models~\citep{team2024chameleon, deng2025emerging, xie2025show} have shown promising results towards this vision, suggesting that truly integrated multimodal intelligence is increasingly within reach.

\begin{figure*}[p]
\centering
\vspace{-4em}
\includegraphics[width=1\linewidth]{assets_our/teaser_v1p1.pdf}
\vspace{-2em}
\caption{We present \model, a native unified multimodal model built on a unified visual representation, enabling diverse multimodal understanding and generation capabilities such as image and video understanding, image and video generation, and image editing.}
\label{fig:teaser}
\vspace{-1em}
\end{figure*}

A central challenge in developing native UMMs lies in how visual inputs are encoded into representations. Current UMMs adopt one of two approaches: (1) decoupled visual representations for understanding and generation tasks, or (2) a unified visual representation shared across both tasks. Intuitively, learning a unified visual representation for both tasks offers compelling advantages for UMMs. 

First, UMMs with decoupled representations, such as BAGEL~\citep{deng2025emerging} and Mogao~\citep{liao2025mogao}, often adopt MoE-style architectures to handle different visual encoders, introducing additional parameters that increase training and inference costs. In contrast, a unified representation allows the model to operate within a single representation space, simplifying training and improving efficiency. Second, different vision encoders typically produce representations with incompatible formats. For the same input, features from a representation encoder (\eg 
SigLIP~\citep{zhai2023sigmoid}) and a causal VAE encoder (\eg Wan 2.1 VAE~\citep{wan2025wan}) differ in (1) spatial compression (16$\times$ vs. 8$\times$), (2) temporal compression (none vs. 4$\times$), and (3) channel dimension (1152 vs. 16). These discrepancies can cause representation conflicts in decoupled models, whereas unified representations inherently avoid such inconsistencies. Finally, unified visual representations provide a clear pathway for achieving mutual enhancement between understanding and generation. While recent studies like Ross~\citep{wang2024reconstructive} and REPA~\citep{yu2024representation} show task-specific improvements in understanding-only and generation-only models, this synergy remains underexplored in existing UMMs.



Nevertheless, current UMMs with unified visual representations often underperform their decoupled counterparts. Most existing approaches adopt a single type of vision encoder for both understanding and generation. For example,  Chameleon~\citep{team2024chameleon} and Transfusion~\citep{zhou2024transfusion} use VQ-VAE~\citep{esser2021taming}, while Harmon~\citep{wu2025harmonizing} utilizes the MAR encoder~\citep{li2024autoregressive}. This unified design tends to favor one task at the expense of the other. Show-o2~\citep{xie2025show} attempts to mitigate this issue by fusing SigLIP~\citep{zhai2023sigmoid} and VAE~\citep{wan2025wan} features through a late-fusion strategy. However, our analysis in Section~\ref{sec:discussion} reveals that its learned representation remains biased toward semantic features, resulting in limited generation quality.

To systematically address these limitations, we propose \model, a native UMM that employs unified visual representations across understanding and generation. Our design is simple yet highly effective: by directly connecting a VAE encoder to a representation encoder, we obtain representations that are sufficiently expressive for diverse multimodal tasks. These unified visual features are fused with text tokens and processed by an LLM decoder, which subsequently generates new text tokens and denoised images through autoregressive next-token prediction and flow matching. As illustrated in \Cref{fig:teaser}, our unified visual representation enables \model to handle image and video understanding, image and video generation, and image editing within a single framework. By conducting a three-stage training, our model achieves state-of-the-art performance on multimodal understanding and generation benchmarks (\eg 61.2\% on MMStar~\citep{chen2024we} and 0.90 on GenEval~\citep{ghosh2023geneval}).

Our main contributions can be summarized as follows:
\begin{itemize}
    \item [1.] We propose \model, a native unified multimodal model with unified visual representations, enabling image/video understanding, image/video generation, and image editing within a single framework.
    \item [2.] Our extensive experiments show that \model's unified visual representation is highly effective, achieving state-of-the-art performance across multiple multimodal understanding and generation tasks.
    \item [3.] We further perform a comprehensive ablation study, demonstrating the superiority of our unified visual representation design over existing methods such as Show-o2 and other models employing decoupled representations.
\end{itemize}

\section{Our Method: \model}
\label{sec:method}
\begin{figure*}[t!]
\centering
\includegraphics[width=1.00\linewidth]{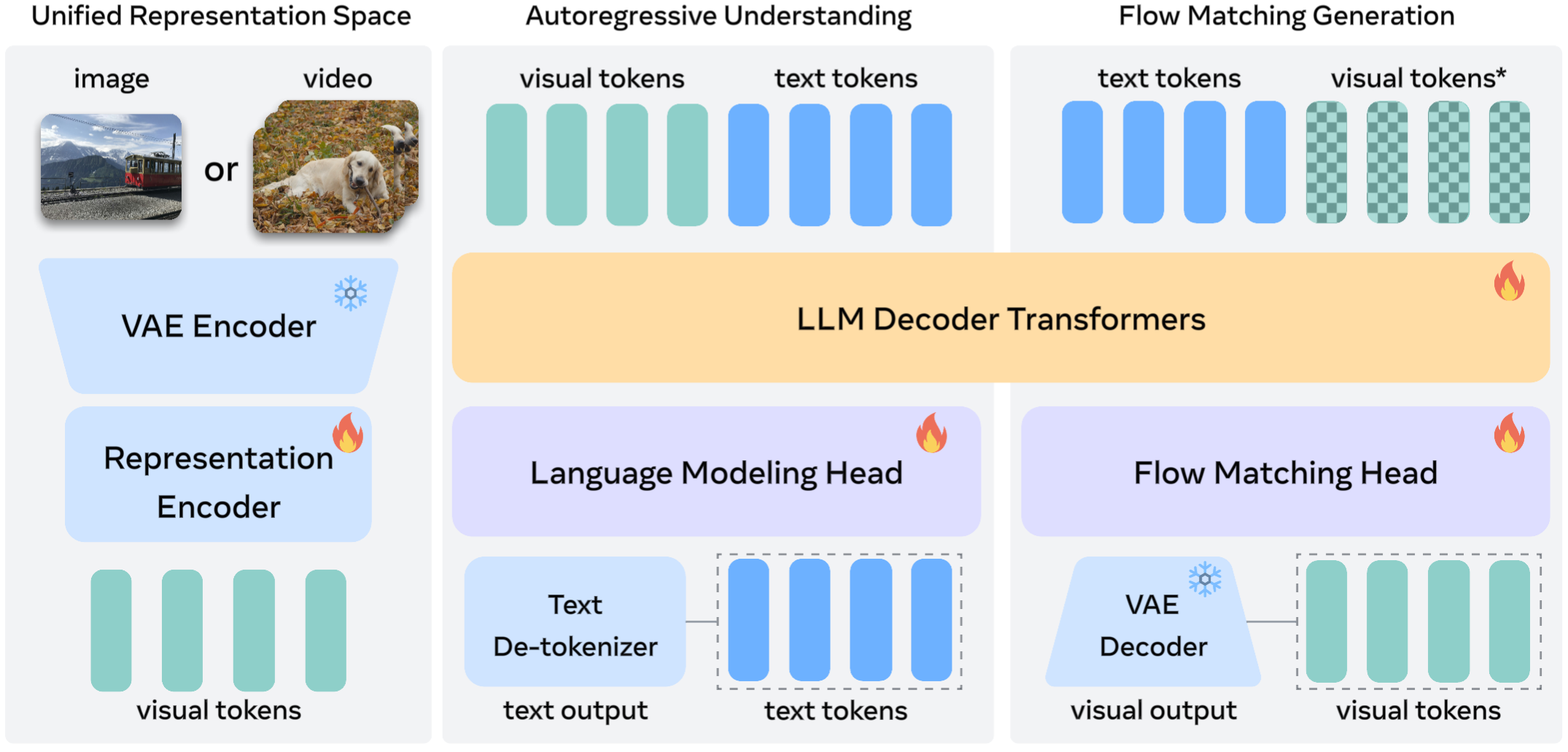}
\vspace{-2em}
\caption{Overview of the \model architecture. Our model employs a VAE encoder and a representation encoder to construct unified visual representations, which are then combined with text tokens and processed by an LLM decoder. The decoder performs autoregressive text generation for understanding tasks and flow-matching-based visual generation for generation tasks. $^*$During visual generation, noise is added to the visual tokens to enable diffusion-based generation.}
\label{fig:framework}
\vspace{-1em}
\end{figure*}

In this section, we introduce \model, a native unified multimodal model that employs unified visual representations across all multimodal understanding and generation tasks. We begin by outlining the key motivations behind our model design in Section~\ref{sec:design_principle}, followed by a detailed description of \model's architecture and training pipeline in Sections~\ref{sec:model_arch} and~\ref{sec:training}, respectively. An overview of our overall framework is shown in Figure~\ref{fig:framework}.

\subsection{Motivation and Design Principles}\label{sec:design_principle}
We discuss the following observations, which motivate the design of \model and its unified visual representations:

\noindent\textbf{Autoregressive vs. diffusion.} Both text and image/video generation can be achieved using autoregressive~\citep{grattafiori2024llama, yang2025qwen3, sun2024autoregressive} or diffusion models~\citep{nie2025large, batifol2025flux, wan2025wan}. In practice, leading understanding-only models~\citep{bai2025qwen2, wang2025internvl3} adopt autoregressive models for text generation. On the other hand, state-of-the-art image and video generators~\citep{esser2024scaling, wan2025wan} employ (latent\footnote{Due to higher generation fidelity and lower training costs, latent diffusion models are generally favoured over pixel diffusion \citep{zheng2025diffusion}.}) diffusion models with flow matching.

\noindent\textbf{Continuous vs. discrete visual representations.} We observe that image and video generation models operating in a continuous (\eg, KL-regularized) VAE latent space~\citep{esser2024scaling, wan2025wan} outperform those using discrete representations~\citep{sun2024autoregressive}, as discretization causes information loss and reduces fidelity. Similarly, multimodal understanding models~\citep{bai2025qwen2, wang2025internvl3} typically rely on continuous semantic features (\eg, CLIP~\citep{radford2021learning} features), suggesting that continuous visual representations are inherently more effective for both understanding and generation tasks.

\noindent\textbf{Semantic representations benefit visual generation.} Recent studies suggest that semantic features enhance visual generation. For instance, REPA~\citep{yu2024representation} demonstrates that diffusion transformers benefit from aligning intermediate features with pretrained representation encoders like DINOv2~\citep{oquab2023dinov2}. Concurrent to our work, RAE~\citep{zheng2025diffusion} employs a frozen representation encoder to encode images into latent representations, showing that pretrained semantic features alone can reconstruct input images effectively.

\noindent\textbf{VAE latents can also support understanding tasks.} We observe that both discrete and continuous VAE latents, originally designed for visual reconstruction, can also support semantic understanding tasks. Recent approaches such as UniTok~\citep{ma2025unitok} and TokLIP~\citep{lin2025toklipmarryvisualtokens} enhance VQ-VAE latents with semantic understanding capability through contrastive learning. Other works explore diffusion models with continuous VAE latents for semantic understanding and dense prediction tasks, including semantic segmentation~\citep{zhu2024unleashing}, object recognition~\citep{li2023your}, and image retrieval~\citep{zuo2024scenediff}.

Building on these observations, we design \model with the following key characteristics:
\begin{itemize}
    \item \model integrates \textbf{autoregressive} text generation with \textbf{flow matching} for image and video generation.
    \item \model builds its unified visual representation on continuous VAE latents, as these latents effectively support both understanding and generation tasks.
    \item To further enhance performance, \model employs \textbf{representation encoders} to extract higher-level features from the VAE latents, improving the quality of both understanding and generation.
\end{itemize}

\subsection{Model Architecture}
\label{sec:model_arch}
\noindent\textbf{Unified visual representations.} As illustrated in \Cref{fig:framework}, \model constructs its unified visual representation using a VAE encoder and a representation encoder. Given an input image or video $\mathbf{X}$, we apply the 3D causal VAE encoder from Wan 2.2~\citep{Wan2.2}, which downsamples the input by 16$\times$ spatially and 4$\times$ temporally, producing the latent $\mathbf{x}_1$. We then generate a noisy latent $\mathbf{x}_t = t\mathbf{x}_1 + (1-t)\mathbf{x}_0$, where $t\in[0, 1]$ is a sampled timestep and $\mathbf{x}_0 \sim \mathcal{N}(0, 1)$. Next, we use the SigLIP 2 vision encoder $\Phi$ (patch size 16, pretrained resolution 512) to extract semantic features from the VAE latents. Since the VAE encoder has 16$\times$ downsampling, we replace SigLIP 2’s original 16$\times$16 patch embedding layer with a randomly initialized 1$\times$1 patch embedding layer, forming a modified encoder $\Phi^\prime$. This ensures that the token sequence lengths of $\Phi(\mathbf{X})$ and $\Phi^\prime(\mathbf{x}_t)$ are consistent. Finally, we apply a two-layer MLP connector to obtain the unified visual representations $\mathbf{z}=\mathtt{MLP}(\Phi^\prime(\mathbf{x}_t))$. During training, we randomly sample $t$ between $[0,1]$ for visual generation and fix $t=1$ for multimodal understanding such that $\mathbf{x}_t$ always corresponds to the clean latent.

For video inputs, where $\mathbf{x}_t\in \mathbb{R}^{b\times c\times f\times h\times w}$ (with $b$ as batch size, $f$ as the number of latent frames, and $c$, $h$, $w$ as channel, height, and width), we aim to prevent the representation encoder $\Phi^\prime$ from processing excessively long sequences. Instead of flattening all latent frames into a single sequence, we apply a window-based attention mechanism by reshaping the frame dimension into the batch dimension in $\Phi^\prime$. In \texttt{einops} notation, the unified visual representation $\mathbf{z}_v$ can be expressed as:
\begin{align}
    \bar{\mathbf{x}}_t & = \mathtt{rearrange}(\mathbf{x}_t, \mathtt{b\ c\ f\ h\ w \rightarrow (b\ f)\ c\ h\ w}),\\
    \bar{\mathbf{z}}_v & =\mathtt{MLP}(\Phi^\prime(\bar{\mathbf{x}}_t)) \in \mathbb{R}^{(b\times f)\times d}, \\
    \mathbf{z}_v & = \mathtt{rearrange}(\bar{\mathbf{z}}_v, \mathtt{(b\ f)\ d \rightarrow b\ (f\ d)}),
\end{align}
where $d$ is the hidden dimension of the video tokens. This operation effectively allows $\Phi^\prime$ to operate independently on each 4-frame window, significantly improving efficiency when processing video tokens.

\begin{figure}[t!]
\centering
\small
\includegraphics[width=0.8\linewidth]{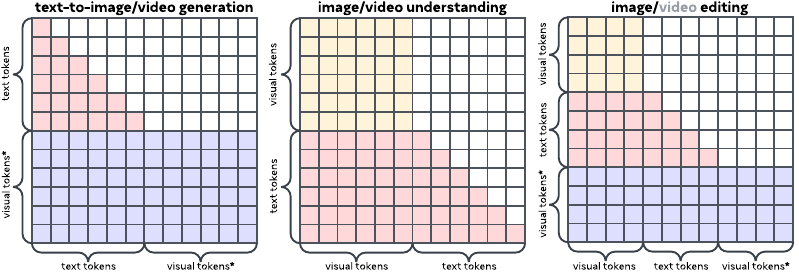}
\vspace{-0.96em}
\caption{Attention masks in the LLM decoder for understanding and generation tasks. $^*$ indicates that the visual tokens are noised.}
\vspace{-1em}
\label{fig:attn_mask}
\end{figure}

\noindent\textbf{LLM decoder and flow matching head.} After obtaining the unified visual representation $\mathbf{z}$, we prepend a timestep token representing the sampled timestep $t$ to $\mathbf{z}$, concatenate this visual token sequence with language tokens and feed the combined sequence into an LLM decoder (Qwen-2.5~\citep{bai2025qwen2}) for joint multimodal processing. Following standard UMM practices~\citep{xie2024show, deng2025emerging}, we apply a causal attention mask on language tokens and a bidirectional attention mask on visual tokens within the LLM decoder layers, as illustrated in \Cref{fig:attn_mask}. For multimodal understanding tasks, the LLM decoder output is passed through a language modeling head to generate text token predictions. For visual generation and image editing, we feed the full token sequence to a randomly initialized flow matching head to predict the velocity for flow matching. This head shares the LLM decoder architecture and adds timestep conditioning via AdaLN-Zero, following Show-o2~\citep{xie2025show} and DiT~\citep{peebles2023scalable}. For generation and editing tasks, we adopt multimodal 3D-RoPE~\citep{seawead2025seaweed, su2024roformer} over the concatenated text-visual sequence to handle interleaved instructions and visual content.

\subsection{Training Pipeline}
\label{sec:training}
To effectively train our unified model, we adopt a three-stage training strategy that progressively adapts each model component to both understanding and generation tasks.

\noindent\textbf{Stage 1: unified representation and flow matching head pretraining.} In the first training stage, our objective is to adapt the semantic representation encoder to generate unified visual representations and to establish a robust initialization for the flow matching head. To this end, we train the representation encoder and flow matching head while freezing the LLM decoder, using two objectives: image captioning and text-to-image generation.

Our image captioning objective aligns with the pretraining objectives of strong semantic encoders, such as SigLIP 2~\citep{tschannen2025siglip} and the Qwen2.5-VL~\citep{bai2025qwen2} vision encoder. Image captioning has also been shown to provide semantic richness comparable to contrastive learning~\citep{tschannen2023image}, thereby enhancing our unified representation's visual understanding capability. Meanwhile, the text-to-image generation objective trains the flow matching head to generate images from text conditions, laying the groundwork for later image editing and text-to-video generation tasks. Additionally, this objective allows generation gradients to flow back into the representation encoder, further aligning our unified visual representation with both understanding and generation tasks.

\noindent\textbf{Stage 2: full model continue pretraining.} In the second training stage, we unfreeze the LLM decoder and pretrain the entire model using the same image captioning and text-to-image generation objectives from Stage 1. During later training steps of Stage 2, we further introduce image instruction-following, image editing, and video-captioning datasets to extend the model’s capabilities. This stage enables \model to perform more complex multimodal reasoning and generation tasks, bridging the gap between basic visual-text alignment and higher-level instruction-driven multimodal understanding and generation.

\noindent\textbf{Stage 3: supervised finetuning (SFT).} Finally, in the third stage, we conduct supervised fine-tuning (SFT) using a combination of image editing, image/video instruction-following, and high-quality image/video generation datasets, trained with a reduced learning rate. This stage further refines \model’s capabilities, improving its performance and generalization across diverse multimodal understanding and generation tasks.
\begin{table*}[t!]
\centering
\scriptsize
\setlength\tabcolsep{4.2 pt}
\begin{tabular}{@{}lccccccccccp{1cm}@{}}
\toprule
\multicolumn{1}{l|}{}                         & \multicolumn{1}{c|}{}                       & \textbf{MME}                           & \textbf{GQA}                         & \textbf{RealWorldQA}    & \textbf{SEED}                                             & \textbf{MMMU}                        & \multicolumn{1}{c}{\textbf{MMStar}}        & \multicolumn{1}{c}{\textbf{AI2D}}   & \multicolumn{1}{c}{\textbf{ChartQA}} & \multicolumn{1}{c}{\textbf{OCRBench}} \\ \cmidrule(l){3-11} 
\multicolumn{1}{l|}{\multirow{-2}{*}{\textbf{Models}}} & \multicolumn{1}{c|}{\multirow{-2}{*}{\textbf{Size}}} & perception                    & test-dev                    & test         & image                                               & val                         & \multicolumn{1}{c}{avg}              &  test                      & \multicolumn{1}{c}{test} & \multicolumn{1}{c}{test} \\ \midrule
\multicolumn{11}{c}{\cellcolor[HTML]{EFEFEF}\textit{Understanding-only Models (LMMs)}}                                                                                                                                                                                                                                       \\
\multicolumn{1}{l|}{LLaVA-1.5 \citep{liu2023visual}}                & \multicolumn{1}{c|}{7B}                       & 1510.7                        & 62.0                        & 54.8      & 65.8                                                & 35.7                           & 33.1                                 & 55.5    & 17.8 & 31.8                           \\
\multicolumn{1}{l|}{Qwen-VL-Chat \citep{bai2023qwen}}             & \multicolumn{1}{c|}{7B}                       & 1487.6                        & 57.5                        & 49.3        & 64.8                                              & 37.0                           & 34.5                                 & 57.7        & 49.8 & 48.8                \\
\multicolumn{1}{l|}{LLaVA-OV \citep{li2024llava}}                 & \multicolumn{1}{c|}{7B}                       & 1580.0                        & -                           & 69.9       & 76.7                                                  & 48.8                        & 57.5                              & 81.4           & 80.9 & 62.2             \\ \midrule
\multicolumn{11}{c}{\cellcolor[HTML]{EFEFEF}\textit{Composite UMMs}}                                                                                                                                                                                                                                                 \\
\multicolumn{1}{l|}{\textcolor{gray}{TokenFlow-XL} \citep{qu2025tokenflow}}                   & \multicolumn{1}{c|}{\textcolor{gray}{14B}}                   & \textcolor{gray}{1551.1}                        & \textcolor{gray}{62.5}                        & \textcolor{gray}{56.6}           & \textcolor{gray}{72.6}                                             & \textcolor{gray}{43.2}                        & \textcolor{gray}{-}                                 & \textcolor{gray}{-}                           & \textcolor{gray}{-}                                & \textcolor{gray}{-} \\ 
\multicolumn{1}{l|}{BLIP3-o \citep{chen2025blip3}}                   & \multicolumn{1}{c|}{4B}                   & 1527.7                        & -                        & 60.4           & 73.8                                              & 46.6                        & -                                 & -                           & -                                & - \\
\multicolumn{1}{l|}{Tar \citep{han2025vision}}                   & \multicolumn{1}{c|}{7B}                   & 1571.0                        & 61.3                        & -               & 73.0                                          & 39.0                        & -                                 & -                           & -                                & - \\
\multicolumn{1}{l|}{X-Omni \citep{geng2025x}}                   & \multicolumn{1}{c|}{7B}                   & -                        & 62.8                        & 62.6           & 74.3                                             & 47.2                        & -                                 & 76.8                           & 81.5                                & 70.4 \\ \midrule
\multicolumn{11}{c}{\cellcolor[HTML]{EFEFEF}\textit{1.5B-scale Native UMMs}}                                                                                                                                                                                                                                                 \\
\multicolumn{1}{l|}{Show-o \citep{xie2024show}}                   & \multicolumn{1}{c|}{1.3B}                   & 1097.2                        & 58.0                        & -       & 51.5                                                 & 27.4                        & -                                 & -                           & -                                & - \\
\multicolumn{1}{l|}{Harmon \citep{wu2025harmonizing}}                   & \multicolumn{1}{c|}{1.5B}                   & 1155.0                        & 58.9                        & 49.8*            & 67.1                                          & \underline{38.9}                        & 35.3*                                 & 57.0*                           & 29.8*                                 & 11.2* \\
\multicolumn{1}{l|}{JanusFlow \citep{ma2025janusflow}}                & \multicolumn{1}{c|}{1.3B}                   & 1333.1                        & \underline{60.3}                        & 41.2*            & \textbf{70.5}                                          & 29.3                        & 40.6*                                 & 54.2                          & \underline{42.4}*                                 & \underline{53.2}* \\
\multicolumn{1}{l|}{SynerGen-VL \citep{li2025synergen}}              & \multicolumn{1}{c|}{2.4B}                   & 1381.0                        & -                           & -              & -                                           & 34.2                        & -                                 & -                           & -                                 & - \\
\multicolumn{1}{l|}{Janus-Pro \citep{chen2025janus}}                & \multicolumn{1}{c|}{1.5B}                   & 1444.0                        & 59.3                        & 52.6*             & 68.3                                         & 36.3                        & 43.1*                                 & 64.5*                           & 23.4                                 & 48.7 \\
\multicolumn{1}{l|}{Show-o2 \citep{xie2025show}}                  & \multicolumn{1}{c|}{1.5B}                   & \underline{1450.9}                        & 60.0                        & \underline{56.5}*             & 65.6                                         & 37.1                        & \underline{43.4}                              & \underline{69.0}                        & 40.0*                                 & 24.5* \\
\rowcolor[HTML]{ECF4FF} 
\multicolumn{1}{l|}{\model}    & \multicolumn{1}{c|}{1.5B}                   &        \textbf{1461.5}                      &    \textbf{61.4}                         &     \textbf{62.5}                 & \underline{69.3}                                   &        \textbf{39.1}                     &        \textbf{54.6}                        &         \textbf{71.4}           &         \textbf{82.1}  &         \textbf{71.9}          \\ \midrule
\multicolumn{11}{c}{\cellcolor[HTML]{EFEFEF}\textit{7B-scale Native UMMs}}                                                                                                                                                                                                                                                   \\
\multicolumn{1}{l|}{\textcolor{gray}{BAGEL} \citep{deng2025emerging}}                     & \multicolumn{1}{c|}{\textcolor{gray}{14B}}                     &  \textcolor{gray}{1687.0}      &  \textcolor{gray}{-} &  \textcolor{gray}{72.8}    & \textcolor{gray}{78.5}     &  \textcolor{gray}{55.3} &  \textcolor{gray}{-}          &  \textcolor{gray}{89.2} & \textcolor{gray}{78.5}                                 & \textcolor{gray}{73.3} \\
\multicolumn{1}{l|}{Emu3 \citep{wang2024emu3}}                     & \multicolumn{1}{c|}{8B}                     &  -      &  60.3 &  57.4    & 68.2     &  31.6 &  -          &  70.0 & -                                 & \underline{68.7} \\
\multicolumn{1}{l|}{VILA-U \citep{wu2024vila}}                   & \multicolumn{1}{c|}{7B}                     &  1401.8 &  60.8 &  -    & 59.0      &  -    &  -          &  -    & -                                 & - \\
\multicolumn{1}{l|}{MUSE-VL \citep{xie2024muse}}                  & \multicolumn{1}{c|}{7B}                     &  -      &  -    &  -    & 69.1    &  39.7 & 49.6       &  69.8 & -                                 & - \\
\multicolumn{1}{l|}{Janus-Pro \citep{chen2025janus}}                & \multicolumn{1}{c|}{7B}                     & 1567.1                        & 62.0                        & 58.0*      & 72.1                                               & 41.0                        & 48.3*                                 & 71.3*                          & 25.8                                 & 59.0  \\
\multicolumn{1}{l|}{Mogao \citep{liao2025mogao}}                    & \multicolumn{1}{c|}{7B}                     & 1592.0                        & 60.9                        & -         & \underline{74.6}                                             & 44.2                        & -                                 & -                          & -                                 & -  \\
\multicolumn{1}{l|}{Show-o2 \citep{xie2025show}}                  & \multicolumn{1}{c|}{7B}                     & \underline{1620.5}                        & \underline{63.1}                        & \underline{64.7}*          & 69.8                                            & \underline{48.9}                        & \underline{56.6}                              & \underline{78.6}          & \underline{52.3}*                                & 32.4*   \\  
\rowcolor[HTML]{ECF4FF} 
\multicolumn{1}{l|}{\model}    & \multicolumn{1}{c|}{7B}                   &        \textbf{1641.5}                      &    \textbf{63.9}                         &     \textbf{66.1}                &             \textbf{74.7}                       &           \textbf{49.8}                  &        \textbf{61.2}                           &         \textbf{79.3}        &        \textbf{85.8}  &         \textbf{74.3}              \\ \bottomrule
\end{tabular}
\vspace{-0.5em}
\caption{Comparisons between \model and baseline models on multimodal understanding benchmarks. Results with model size greater than 13B are \textcolor{gray}{grayed}. \textbf{Bold}: best results among each section. \underline{Underline}: second-best. * indicates the results based on our evaluation scripts.}
\label{tab:img_und}
\vspace{-1em}
\end{table*}

\section{Experiments}
\label{sec:exp}

\subsection{Experiment Setup}

\noindent\textbf{Implementation details.} 
We verify \model with two LLM models at different scales, \ie, Qwen2.5-1.5B-Instruct and Qwen2.5-7B-Instruct~\citep{bai2025qwen2}. In the pretraining stage, we optimize the representation encoder, projection layers, and diffusion head with AdamW~\citep{loshchilov2017decoupled} using a learning rate of $1\times10^{-4}$. We train on images with a base resolution of 512$\times$512, as well as alternative aspect ratios that yield a similar number of visual tokens. In the second stage, we enable end-to-end training after a linear warm-up of 2K steps and continue optimization with the same learning rate. At this point, we extend the training data to include video-caption pairs and editing data. In the final stage, we perform supervised fine-tuning for instruction following on our curated SFT corpus with a smaller learning rate of $2\times10^{-5}$. Due to the substantial computational cost of video training, the 7B variant is trained without video data.

\begin{table}[!t]
\centering
\small
\setlength\tabcolsep{4.2 pt}
\begin{tabular}{@{}lccccccccp{1cm}@{}}
\toprule
\multicolumn{1}{l|}{\textbf{Models}}                                        & \multicolumn{1}{c|}{\textbf{Size}}                         & \textbf{Single Obj.} & \textbf{Two Obj.} & \textbf{Counting} & \textbf{Colors} & \textbf{Position} & \multicolumn{1}{c|}{\textbf{Color Attr.}}              & \multicolumn{1}{c}{\textbf{Overall}} \\ \midrule
\rowcolor[HTML]{EFEFEF} 
\multicolumn{9}{c}{\cellcolor[HTML]{EFEFEF}\textit{Generation-only Models}}                                                                                                                                                                                   \\
\multicolumn{1}{l|}{SD3-Medium \citep{esser2024scaling}}                                    & \multicolumn{1}{c|}{2B}                            & 0.99        & 0.94     & 0.72     & 0.89   & 0.33     & \multicolumn{1}{c|}{0.60}                      & 0.74                        \\ 
\multicolumn{1}{l|}{FLUX.1 [Dev]$^{\dagger}$ \citep{batifol2025flux}}                                    & \multicolumn{1}{c|}{12B}                            & 0.98        & 0.93   & 0.75     & 0.93   & 0.68     & \multicolumn{1}{c|}{0.65}                      & 0.82                      \\ \midrule
\rowcolor[HTML]{EFEFEF} 

\multicolumn{9}{c}{\cellcolor[HTML]{EFEFEF}\textit{Composite UMMs}}                                                                                                                                                                    \\
\multicolumn{1}{l|}{MetaQuery-XL$^{\dagger}$\citep{pan2025transfer}}                                       & \multicolumn{1}{c|}{7B}                        & -      & -      & -         & -        & -                         & \multicolumn{1}{c|}{-}       & 0.80                   \\
\multicolumn{1}{l|}{Tar \citep{han2025vision}}                                       & \multicolumn{1}{c|}{7B}                         & 0.99  & 0.92  & 0.83     & 0.85    & 0.80                     & \multicolumn{1}{c|}{0.65 }            & 0.84                \\

\multicolumn{1}{l|}{BLIP3-o \citep{chen2025blip3}}                                       & \multicolumn{1}{c|}{8B}                         & -      & -      & -         & -        & -                         & \multicolumn{1}{c|}{-}       & 0.84                    \\
\multicolumn{1}{l|}{UniWorld-V1$^{\dagger}$ \citep{lin2025uniworld}}                                       & \multicolumn{1}{c|}{12B}                        & 0.98      & 0.93     & 0.81         & 0.89        & 0.74                         & \multicolumn{1}{c|}{0.71}       & 0.84                   \\ 
\multicolumn{1}{l|}{OmniGen2$^{\dagger}$ \citep{wu2025omnigen2}}                                       & \multicolumn{1}{c|}{7B}                        & 0.99      & 0.96     & 0.74         & 0.98        & 0.71                         & \multicolumn{1}{c|}{0.75}       & 0.86                   \\
                         \midrule
                         
\multicolumn{9}{c}{\cellcolor[HTML]{EFEFEF}\textit{1.5B-scale Native UMMs}}                                                                                                                                                                                   \\
\multicolumn{1}{l|}{D-DiT \citep{li2025dual}}                                         & \multicolumn{1}{c|}{2B}                           & 0.97        & 0.80     & 0.54     & 0.76   & 0.32     & \multicolumn{1}{c|}{0.50}                      & 0.65                        \\
\multicolumn{1}{l|}{Show-o \citep{xie2024show}}                                        & \multicolumn{1}{c|}{1.5B}                         & 0.98        & 0.80     & \underline{0.66}     & 0.84   & 0.31     & \multicolumn{1}{c|}{0.50}                      & 0.68                        \\
\multicolumn{1}{l|}{Janus-Pro \citep{chen2025janus}}                                     & \multicolumn{1}{c|}{1.5B}                         & 0.98        & 0.82     & 0.51     & 0.89   & 0.65     & \multicolumn{1}{c|}{0.56}                      & 0.73                        \\
\multicolumn{1}{l|}{Show-o2 \citep{xie2025show}}                                       & \multicolumn{1}{c|}{1.5B}                         & \underline{0.99}        & \underline{0.86}     & 0.55     & \underline{0.86}   & 0.46     & \multicolumn{1}{c|}{\underline{0.63}}                      & 0.73                        \\
\multicolumn{1}{l|}{Harmon \citep{wu2025harmonizing}}                                        & \multicolumn{1}{c|}{1.5B}                         & \underline{0.99}        & \underline{0.86}     & \underline{0.66}     & 0.85   & \underline{0.74}     & \multicolumn{1}{c|}{0.48}                      & \underline{0.76}                        \\
\rowcolor[HTML]{ECF4FF} 
\multicolumn{1}{l|}{\cellcolor[HTML]{ECF4FF}\model} & \multicolumn{1}{c|}{\cellcolor[HTML]{ECF4FF}1.5B} & \textbf{1.00}           & \textbf{0.94}        & \textbf{0.83}        & \textbf{0.91}      & \textbf{0.81}        & \multicolumn{1}{c|}{\cellcolor[HTML]{ECF4FF}\textbf{0.79}} & \textbf{0.88}                           \\ \midrule
\rowcolor[HTML]{EFEFEF} 
\multicolumn{9}{c}{\cellcolor[HTML]{EFEFEF}\textit{7B-scale Native UMMs}}                                                                                                                                                                                     \\
\multicolumn{1}{l|}{MUSE-VL \citep{xie2025muse}}                                       & \multicolumn{1}{c|}{7B}                           & -           & -        & -        & -      & -        & \multicolumn{1}{c|}{-}                         & 0.57                        \\
\multicolumn{1}{l|}{Transfusion \citep{zhou2024transfusion}}                                   & \multicolumn{1}{c|}{7B}                           & -           & -        & -        & -      & -        & \multicolumn{1}{c|}{-}                         & 0.63                        \\
\multicolumn{1}{l|}{Emu3 \citep{wang2024emu3}}                                          & \multicolumn{1}{c|}{8B}                           & -           & -        & -        & -      & -        & \multicolumn{1}{c|}{-}                         & 0.66                        \\
\multicolumn{1}{l|}{Show-o2 \citep{xie2025show}}                                       & \multicolumn{1}{c|}{7B}                           & \textbf{1.00}        & 0.87     & 0.58     & 0.92   & 0.52     & \multicolumn{1}{c|}{0.62}                      & 0.76                        \\
\multicolumn{1}{l|}{Janus-Pro \citep{chen2025janus}}                                     & \multicolumn{1}{c|}{7B}                           & \underline{0.99}        & 0.89     & 0.59     & 0.90   & 0.79     & \multicolumn{1}{c|}{0.66}                      & 0.80                        \\
\multicolumn{1}{l|}{BAGEL$^{\dagger}$ \citep{deng2025emerging}}                                         & \multicolumn{1}{c|}{14B}                          & 0.98        & \underline{0.95}     & \textbf{0.84}     & \textbf{0.95}   & 0.78     & \multicolumn{1}{c|}{0.77}                      & 0.88                        \\
\multicolumn{1}{l|}{Mogao \citep{liao2025mogao}}                                         & \multicolumn{1}{c|}{7B}                           & \textbf{1.00}        & \textbf{0.97}     & \underline{0.83}     & \underline{0.93}   & \underline{0.84}     & \multicolumn{1}{c|}{\underline{0.80}}                      & \underline{0.89}                        \\
\rowcolor[HTML]{ECF4FF} 
\multicolumn{1}{l|}{\cellcolor[HTML]{ECF4FF}\model} & \multicolumn{1}{c|}{\cellcolor[HTML]{ECF4FF}7B}   & \textbf{1.00}           & \textbf{0.97}        &    0.81     &    0.91   & \textbf{0.88}        & \multicolumn{1}{c|}{\cellcolor[HTML]{ECF4FF}\textbf{0.83}} & \textbf{0.90}                   \\ \bottomrule
\end{tabular}
\vspace{-0.5em}
\caption{Image generation results on GenEval. $^\dagger$ refers to methods using LLM rewriters. \textbf{Bold}: best results among each section. \underline{Underline}: second-best.}
\label{tab:gen_eval}
\vspace{-1em}
\end{table}

\subsection{Main Results}
\label{sec:eval}

\noindent\textbf{Image understanding.} We evaluate \model's multimodal understanding capabilities on nine benchmarks, including general VQA benchmarks such as MME~\citep{fu2025mmecomprehensiveevaluationbenchmark}, GQA~\citep{hudson2019gqa}, RealWorldQA~\citep{xai2024-grok15v} and SEED-Bench~\citep{li2023seed}; knowledge-intensive benchmarks such as MMMU~\citep{yue2024mmmu}, MMStar~\citep{chen2024we}, and AI2D~\citep{kembhavi2016diagram}; and text-centric benchmarks including ChartQA~\citep{masry2022chartqa} and OCRBench~\citep{liu2024ocrbench}. As shown in Table~\ref{tab:img_und}, both 1.5B and 7B \model achieve state-of-the-art results across nearly all benchmarks, demonstrating strong and consistent performance. Notably, \model delivers competitive image understanding results compared to understanding-only models and outperforms many composite UMMs and UMMs with larger model sizes, highlighting the effectiveness of its unified representations.

\noindent\textbf{Image generation.}
We evaluate \model's image generation performance on three benchmarks: GenEval~\citep{ghosh2023geneval}, DPG-Bench~\citep{hu2024ella} and OneIG-Bench~\citep{chang2025oneig}. Results are presented in Table~\ref{tab:gen_eval} and Table~\ref{tab:dpg_bench}. Across all three benchmarks, \model consistently outperforms contemporary approaches such as Janus-Pro, BAGEL and Mogao, achieving state-of-the-art results for both the 1.5B and 7B variants. Notably, \model shows a substantial advantage in text rendering quality in OneIG-Bench, indicating its strong semantic understanding capability when generating images from complex instructions containing visual text-related information. Our results show that \model consistently outperforms models with decoupled visual representations on image generation tasks, underscoring the strength and robustness of its unified representation design.

\begin{table}[!t]
\centering
\scriptsize
\setlength\tabcolsep{2.2 pt}
\begin{tabular}{@{}lcccccccccccccp{1cm}@{}}
\toprule
\multicolumn{1}{l|}{}                                              & \multicolumn{1}{l|}{}                                & \multicolumn{6}{c|}{\textbf{DPG-Bench}}                                                                                                                                                                                                                                           & \multicolumn{6}{c}{\textbf{OneIG-Bench}}                                                                                                              \\ \cmidrule(l){3-14} 
\multicolumn{1}{l|}{\multirow{-2}{*}{\textbf{Models}}}             & \multicolumn{1}{c|}{\multirow{-2}{*}{\textbf{Size}}} & \textbf{Global}                        & \textbf{Entity}                     & \textbf{Attribute}                  & \textbf{Relation}                   & \multicolumn{1}{c|}{\textbf{Other}}                      & \multicolumn{1}{c|}{\textbf{Overall}}                       & \textbf{Alignment} & \textbf{Text} & \textbf{Reasoning} & \textbf{Style} & \multicolumn{1}{c|}{\textbf{Diversity}}                 & \textbf{Average} \\ \midrule
\rowcolor[HTML]{EFEFEF} 
\multicolumn{14}{c}{\cellcolor[HTML]{EFEFEF}\textit{Generation-only Models}}                                                                                                                                                                                                                                                                                                                                                                                                                                                                                          \\
\multicolumn{1}{l|}{FLUX.1 {[}Dev{]} \citep{batifol2025flux}}                              & \multicolumn{1}{l|}{12B}                             & 82.10                                  & 89.50                               & 88.70                               & 91.10                               & \multicolumn{1}{c|}{89.40}                               & \multicolumn{1}{c|}{84.00}                                  & 0.79               & 0.52          & 0.25               & 0.37           & \multicolumn{1}{c|}{0.24}                               & 0.43             \\
\multicolumn{1}{l|}{Qwen-Image \citep{wu2025qwen}}                                    & \multicolumn{1}{l|}{20B}                             & 91.32                                  & 91.56                               & 92.02                               & 94.31                               & \multicolumn{1}{c|}{92.73}                               & \multicolumn{1}{c|}{88.32}                                  & 0.88               & 0.89          & 0.31               & 0.42           & \multicolumn{1}{c|}{0.20}                               & 0.54             \\ \midrule
\rowcolor[HTML]{EFEFEF} 
\multicolumn{14}{c}{\cellcolor[HTML]{EFEFEF}\textit{1.5B-scale Native UMMs}}                                                                                                                                                                                                                                                                                                                                                                                                                                                                                          \\
\multicolumn{1}{l|}{Show-o \citep{xie2024show}}                                        & \multicolumn{1}{l|}{1.3B}                            & -                                      & -                                   & -                                   & -                                   & \multicolumn{1}{c|}{-}                                   & \multicolumn{1}{c|}{-}                                      & 0.70               & 0.00          & 0.21               & \textbf{0.36}  & \multicolumn{1}{c|}{\textbf{0.24}}                      & 0.25             \\
\multicolumn{1}{l|}{Show-o2 \citep{xie2025show}}                                       & \multicolumn{1}{c|}{1.5B}                            & {\ul 87.53}                            & \textbf{90.38}                      & {\ul 91.34}                         & {\ul 90.30}                         & \multicolumn{1}{c|}{\textbf{91.21}}                      & \multicolumn{1}{c|}{{\ul 85.02}}                            & {\ul 0.80}         & {\ul 0.13}    & \textbf{0.27}      & {\ul 0.35}     & \multicolumn{1}{c|}{0.19}                               & {\ul 0.35}       \\
\rowcolor[HTML]{ECF4FF} 
\multicolumn{1}{l|}{\cellcolor[HTML]{ECF4FF}\model} & \multicolumn{1}{c|}{\cellcolor[HTML]{ECF4FF}1.5B}    & \textbf{88.87}                         & {\ul 90.32}                         & \textbf{91.71}                      & \textbf{91.79}                      & \multicolumn{1}{c|}{\cellcolor[HTML]{ECF4FF}{\ul 90.14}} & \multicolumn{1}{c|}{\cellcolor[HTML]{ECF4FF}\textbf{86.03}} & \textbf{0.82}      & \textbf{0.77} & {\ul 0.25}         & \textbf{0.36}  & \multicolumn{1}{c|}{\cellcolor[HTML]{ECF4FF}{\ul 0.20}} & \textbf{0.48}    \\ \midrule
\rowcolor[HTML]{EFEFEF} 
\multicolumn{14}{c}{\cellcolor[HTML]{EFEFEF}\textit{7B-scale Native UMMs}}                                                                                                                                                                                                                                                                                                                                                                                                                                                                                            \\
\multicolumn{1}{l|}{Emu3-DPO \citep{wang2024emu3}}                                      & \multicolumn{1}{c|}{8B}                              & -                                      & -                                   & -                                   & -                                   & \multicolumn{1}{c|}{-}                                   & \multicolumn{1}{c|}{81.60}                                  & -                  & -             & -                  & -              & \multicolumn{1}{c|}{-}                                  & -                \\
\multicolumn{1}{l|}{Janus-Pro \citep{chen2025janus}}                                     & \multicolumn{1}{c|}{7B}                              & 86.90                                  & 88.90                               & 89.40                               & 89.32                               & \multicolumn{1}{c|}{89.48}                               & \multicolumn{1}{c|}{84.19}                                  & 0.55               & 0.00          & 0.14               & 0.28           & \multicolumn{1}{c|}{\textbf{0.37}}                      & 0.27             \\
\multicolumn{1}{l|}{Mogao \citep{liao2025mogao}}                                         & \multicolumn{1}{c|}{7B}                              & 82.37                                  & 90.03                               & 88.26                               & \textbf{93.18}                      & \multicolumn{1}{c|}{85.40}                               & \multicolumn{1}{c|}{84.33}                                  & -                  & -             & -                  & -              & \multicolumn{1}{c|}{-}                                  & -                \\
\multicolumn{1}{l|}{BAGEL \citep{deng2025emerging}}                                         & \multicolumn{1}{l|}{14B}                             & 88.94                                  & 90.37                               & \textbf{91.29}                      & 90.82                               & \multicolumn{1}{c|}{88.67}                               & \multicolumn{1}{c|}{85.07}                                  & 0.77               & {\ul 0.24}    & 0.17               & {\ul 0.37}     & \multicolumn{1}{c|}{{\ul 0.25}}                         & {\ul 0.36}       \\
\multicolumn{1}{l|}{Show-o2 \citep{xie2025show}}                                       & \multicolumn{1}{c|}{7B}                              & {\ul 89.00}                            & \textbf{91.78}                      & 89.96                               & 91.81                               & \multicolumn{1}{c|}{\textbf{91.64}}                      & \multicolumn{1}{c|}{{\ul 86.14}}                            & {\ul 0.82}         & 0.00          & {\ul 0.23}         & 0.32           & \multicolumn{1}{c|}{0.18}                               & 0.31             \\
\rowcolor[HTML]{ECF4FF} 
\multicolumn{1}{l|}{\cellcolor[HTML]{ECF4FF}\model} & \multicolumn{1}{c|}{\cellcolor[HTML]{ECF4FF}7B}      & \cellcolor[HTML]{ECF4FF}\textbf{90.42} & \cellcolor[HTML]{ECF4FF}{\ul 91.68} & \cellcolor[HTML]{ECF4FF}{\ul 90.94} & \cellcolor[HTML]{ECF4FF}{\ul 91.87} & \multicolumn{1}{c|}{\cellcolor[HTML]{ECF4FF}{\ul 90.73}} & \multicolumn{1}{c|}{\cellcolor[HTML]{ECF4FF}\textbf{86.76}} & \textbf{0.84}      & \textbf{0.82} & \textbf{0.27}      & \textbf{0.40}  & \multicolumn{1}{c|}{\cellcolor[HTML]{ECF4FF}0.19}       & \textbf{0.50}    \\ \bottomrule
\end{tabular}
\vspace{-0.5em}
\caption{Image generation results on DPG-Bench. \textbf{Bold}: best results among each section. \underline{Underline}: second-best.}
\label{tab:dpg_bench}
\end{table}

\begin{table}[!t]
\centering
\scriptsize
\setlength\tabcolsep{3 pt}
\begin{tabular}{@{}lccccccccccccccp{1cm}@{}}
\toprule
\multicolumn{1}{l|}{}                                              & \multicolumn{1}{l|}{}                                & \multicolumn{10}{c|}{\textbf{ImgEdit-Bench}}                                                                                                                                                                                                                                                                                                                                    & \multicolumn{3}{c}{\textbf{GEdit-Bench}}                                                        \\ \cmidrule(l){3-15} 
\multicolumn{1}{l|}{\multirow{-2}{*}{\textbf{Models}}}             & \multicolumn{1}{c|}{\multirow{-2}{*}{\textbf{Size}}} & \textbf{Add}                          & \textbf{Adj.}                         & \textbf{Ext.}                         & \textbf{Rep.}                         & \textbf{Rm.}                          & \textbf{Bg.}  & \textbf{Sty.} & \textbf{Hyb.} & \multicolumn{1}{c|}{\textbf{Act.}}                         & \multicolumn{1}{c|}{\textbf{Overall}}                      & \textbf{G-SC} & \multicolumn{1}{c|}{\textbf{G-PQ}}                         & \textbf{G-Overall} \\ \midrule
\rowcolor[HTML]{EFEFEF} 
\multicolumn{15}{c}{\cellcolor[HTML]{EFEFEF}\textit{Generation-only Models}}                                                                                                                                                                                                                                                                                                                                                                                                                                                                                                                                  \\
\multicolumn{1}{l|}{FLUX.1 Kontext {[}Pro{]} \citep{batifol2025flux}}                      & \multicolumn{1}{l|}{12B}                             & 4.25                                  & 4.15                                  & 2.35                                  & 4.56                                  & 3.57                                  & 4.26          & 4.57          & 3.68          & \multicolumn{1}{c|}{4.63}                                  & \multicolumn{1}{c|}{4.00}                                  & 7.02          & \multicolumn{1}{c|}{7.60}                                  & 6.56               \\
\multicolumn{1}{l|}{Qwen-Image \citep{wu2025qwen}}                                    & \multicolumn{1}{l|}{20B}                             & 4.38                                  & 4.16                                  & 3.43                                  & 4.66                                  & 4.14                                  & 4.38          & 4.81          & 3.82          & \multicolumn{1}{c|}{4.69}                                  & \multicolumn{1}{c|}{4.27}                                  & 8.00          & \multicolumn{1}{c|}{7.86}                                  & 7.56               \\ \midrule
\rowcolor[HTML]{EFEFEF} 
\multicolumn{15}{c}{\cellcolor[HTML]{EFEFEF}\textit{Native or Composite UMMs}}                                                                                                                                                                                                                                                                                                                                                                                                                                                                                                                                \\
\multicolumn{1}{l|}{OmniGen \citep{xiao2025omnigen}}                                       & \multicolumn{1}{c|}{3.8B}                            & 3.47                                  & 3.04                                  & 1.71                                  & 2.94                                  & 2.43                                  & 3.21          & 4.19          & 2.24          & \multicolumn{1}{c|}{3.38}                                  & \multicolumn{1}{c|}{2.96}                                  & 5.96          & \multicolumn{1}{c|}{5.89}                                  & 5.06               \\
\multicolumn{1}{l|}{BAGEL \citep{deng2025emerging}}                                         & \multicolumn{1}{c|}{14B}                             & 3.56                                  & 3.31                                  & 1.70                                  & 3.30                                  & 2.62                                  & 3.24          & 4.49          & 2.38          & \multicolumn{1}{c|}{4.17}                                  & \multicolumn{1}{c|}{3.20}                                  & {\ul 7.36}    & \multicolumn{1}{c|}{6.83}                                  & {\ul 6.52}         \\
\multicolumn{1}{l|}{UniWorld-V1 \citep{lin2025uniworld}}                                   & \multicolumn{1}{c|}{12B}                             & {\ul 3.82}                            & {\ul 3.64}                            & {\ul 2.27}                            & 3.47                                  & {\ul 3.24}                            & 2.99          & 4.21          & {\ul 2.96}    & \multicolumn{1}{c|}{2.74}                                  & \multicolumn{1}{c|}{3.26}                                  & 4.93          & \multicolumn{1}{c|}{{\ul 7.43}}                            & 4.85               \\
\multicolumn{1}{l|}{OmniGen2 \citep{wu2025omnigen2}}                                      & \multicolumn{1}{c|}{4B}                              & 3.57                                  & 3.06                                  & 1.77                                  & {\ul 3.74}                            & 3.20                                  & {\ul 3.57}    & \textbf{4.81} & 2.52          & \multicolumn{1}{c|}{{\ul 4.68}}                            & \multicolumn{1}{c|}{{\ul 3.44}}                            & 7.16          & \multicolumn{1}{c|}{6.77}                                  & 6.41               \\
\rowcolor[HTML]{ECF4FF} 
\multicolumn{1}{l|}{\cellcolor[HTML]{ECF4FF}\model} & \multicolumn{1}{c|}{\cellcolor[HTML]{ECF4FF}7B}      & \cellcolor[HTML]{ECF4FF}\textbf{4.46} & \cellcolor[HTML]{ECF4FF}\textbf{4.52} & \cellcolor[HTML]{ECF4FF}\textbf{2.47} & \cellcolor[HTML]{ECF4FF}\textbf{4.68} & \cellcolor[HTML]{ECF4FF}\textbf{4.58} & \textbf{4.56} & {\ul 4.73}    & \textbf{4.07} & \multicolumn{1}{c|}{\cellcolor[HTML]{ECF4FF}\textbf{4.69}} & \multicolumn{1}{c|}{\cellcolor[HTML]{ECF4FF}\textbf{4.31}} & \textbf{7.79} & \multicolumn{1}{c|}{\cellcolor[HTML]{ECF4FF}\textbf{7.48}} & \textbf{7.29}      \\ \bottomrule
\end{tabular}
\vspace{-0.5em}
\caption{Image editing results on ImgEdit-Bench and GEdit-Bench. For ImgEdit-Bench, we test editing performance across various dimensions, including `Add', `Adjust', `Extract', `Replace', `Remove', `Background', `Style', `Hybrid', and `Action'. For GEdit-Bench, ``G-SC'' and ``G-PQ'' denote ``G-Semantic Consistency'' and ``G-Perceptual Quality'', respectively. \textbf{Bold}: best results among each section. \underline{Underline}: second-best.}
\label{tab:img_edit}
\vspace{-1em}
\end{table}

\noindent\textbf{Image editing.}
We employ ImgEdit-Bench~\citep{ye2025imgedit} and GEdit-Bench as our evaluation suite for image editing. As shown in Table~\ref{tab:img_edit}, \model achieves an overall score of 4.31 on ImgEdit-Bench, ranking highest among all UMMs. \model's performance is also comparable to generation-only models such as FLUX.1 Kontext~\citep{batifol2025flux} and Qwen-Image~\citep{wu2025qwen}. For GEdit-Bench, although \model performs slightly below the best generation-only model (Qwen-Image~\citep{wu2025qwen}), it again achieves the highest overall score among all unified models. \model's consistently strong results on both ImgEdit-Bench and GEdit-Bench demonstrate its robust image editing capability and highlight the effectiveness of our unified visual representation when handling visual generation tasks that require precise semantic understanding and accurate prompt following.

\noindent\textbf{Video understanding.} We employ four video understanding benchmarks to evaluate \model: MVBench~\citep{li2024mvbench}, Video-MME~\citep{fu2025video}, LongVideoBench~\citep{wu2024longvideobench} and LVBench~\citep{wang2025lvbench}. As shown in Table~\ref{tab:vid_und}, \model outperforms Show-o2 on MVBench and Video-MME, while achieving competitive results on LongVideoBench and LVBench. Notably, despite being only a 1.5B-parameter model, \model performs on par with larger understanding-only models on MVBench and LVBench, demonstrating the efficiency and effectiveness of our unified representation for video understanding tasks.

\noindent\textbf{Video generation.}
We evaluate \model on VBench~\citep{huang2024vbench} for text-to-video generation, comparing it against other UMMs and generation-only models. As shown in Table~\ref{tab:vbench}, \model achieves state-of-the-art performance, surpassing all existing UMMs capable of video generation, while using only a 1.5B-parameter LLM decoder. This demonstrates the efficiency and scalability of our unified architecture for high-quality video generation.

\begin{table}[!t]
\centering
\small

\setlength\tabcolsep{4 pt}
\begin{tabular}{@{}lccccccp{1cm}@{}}
\toprule
\multicolumn{1}{l|}{}                         & \multicolumn{1}{c|}{}  & \multicolumn{1}{c|}{}                     & \textbf{MVBench}   & \multicolumn{1}{c}{\textbf{Video-MME}}      & \textbf{LongVideoBench} &  \textbf{LVBench} \\ \cmidrule(l){4-7} 
\multicolumn{1}{l|}{\multirow{-2}{*}{\textbf{Models}}} & \multicolumn{1}{c|}{\multirow{-2}{*}{\textbf{Size}}} & \multicolumn{1}{c|}{\multirow{-2}{*}{\textbf{\#Frames}}} & test      & \multicolumn{1}{c}{w/o sub }      & val & test           \\ \midrule

\multicolumn{7}{c}{\cellcolor[HTML]{EFEFEF}\textit{Understanding-only Models (LMMs)}} \\

\multicolumn{1}{l|}{GPT-4o \citep{OpenAI_GPT4o}}                   
& \multicolumn{1}{c|}{-} & \multicolumn{1}{c|}{-}                      
& -         
& \multicolumn{1}{c}{71.9}    
& 66.7   
& 48.9        \\

\multicolumn{1}{l|}{Gemini-1.5-Pro \citep{team2024gemini}}           
& \multicolumn{1}{c|}{-} & \multicolumn{1}{c|}{-}                      
& 54.2      
& \multicolumn{1}{c}{75.0}    
& 64.0  
& 33.1         \\

\multicolumn{1}{l|}{LongVA \citep{zhang2024long}}                   
& \multicolumn{1}{c|}{7B} & \multicolumn{1}{c|}{64}                    
& 49.2      
& \multicolumn{1}{c}{52.6}    
& 51.8   
& -        \\

\multicolumn{1}{l|}{VideoLLaMA2 \citep{cheng2024videollama}}              
& \multicolumn{1}{c|}{7B}   & \multicolumn{1}{c|}{16}                  
& 54.6      
& \multicolumn{1}{c}{47.9}       
& -     
& -         \\

\multicolumn{1}{l|}{LLaVA-OV \citep{li2024llava}}                 
& \multicolumn{1}{c|}{7B}           & \multicolumn{1}{c|}{32}          
& 56.7      
& \multicolumn{1}{c}{58.2}    
& 56.5  
& 26.9         \\ \midrule

\multicolumn{7}{c}{\cellcolor[HTML]{EFEFEF}\textit{1.5B-scale Native UMMs}} \\

\multicolumn{1}{l|}{Show-o2 \citep{xie2025show}}                  
& \multicolumn{1}{c|}{1.5B}       & \multicolumn{1}{c|}{32}            
& \underline{49.8}      
& \multicolumn{1}{c}{\underline{48.0}}    
& \underline{49.2}  
& -       \\ 

\rowcolor[HTML]{ECF4FF} 
\multicolumn{1}{l|}{\model}    
& \multicolumn{1}{c|}{1.5B}    & \multicolumn{1}{c|}{49}               
& \textbf{54.4}       
& \multicolumn{1}{c}{\textbf{49.1}}          
& \textbf{49.7} 
& \textbf{27.4}  \\ 

\bottomrule
\end{tabular}

\vspace{-0.5em}
\caption{Experimental results on video understanding benchmarks. \#Frames denotes the number of frames used during inference. \textbf{Bold}: best results. \underline{Underline}: second-best.}
\label{tab:vid_und}
\end{table}

\begin{table*}[!t]
\centering
\scriptsize
\setlength\tabcolsep{1.65 pt}
\begin{tabular}{@{}lccccccccccccccccccccp{1cm}@{}}
\toprule
\multicolumn{1}{l|}{\textbf{Models}}                                        & \multicolumn{1}{c|}{\textbf{Size}}                         & \textbf{QS}    & \textbf{SS}    & \textbf{SC}    & \textbf{BC}    & \textbf{TF}    & \textbf{MS}    & \textbf{DD}    & \textbf{AQ}    & \textbf{IQ}    & \textbf{OC}    & \textbf{MO}    & \textbf{HA}    & \textbf{C}     & \textbf{SR}    & \textbf{S}     & \textbf{AS}    & \textbf{TS}    & \multicolumn{1}{c|}{\textbf{OC'}}                       & \textbf{Total} \\ \midrule
\rowcolor[HTML]{EFEFEF} 
\multicolumn{21}{c}{\cellcolor[HTML]{EFEFEF}\textit{Generation-only Models}}                                                                                                                                                                                                                                            \\
\multicolumn{1}{l|}{CogVideoX}                                     & \multicolumn{1}{c|}{5B}                           & 82.75 & 77.04 & 96.23 & 96.52 & 98.66 & 96.92 & 70.97 & 61.98 & 62.90 & 85.23 & 62.11 & 99.40 & 82.81 & 66.35 & 53.20 & 24.91 & 25.38 & \multicolumn{1}{c|}{27.59}                     & 81.61 \\ \midrule
\rowcolor[HTML]{EFEFEF} 
\multicolumn{21}{c}{\cellcolor[HTML]{EFEFEF}\textit{Native or Composite UMMs}}                                                                                                                                                                                                                                                       \\
\multicolumn{1}{l|}{VILA-U}                                       & \multicolumn{1}{c|}{7B}                           & 76.26 & 65.04 & -     & -     & -     & -     & -     & -     & -     & -     & -     & -     & -     & -     & -     & -     & -     & \multicolumn{1}{c|}{-}                         & 74.01 \\
\multicolumn{1}{l|}{HaploOmni}                                     & \multicolumn{1}{c|}{7B}                           & -     & -     & \underline{96.40} & \underline{97.60} & -     & 96.80 & 65.30 & -     & -     & -     & -     & -     & -     & -     & 34.60 & -     & -     & \multicolumn{1}{c|}{-}                         & 78.10 \\
\multicolumn{1}{l|}{Emu3}                                          & \multicolumn{1}{c|}{8B}                           & -     & -     & 95.32 & \textbf{97.69} & -     & \textbf{98.93} & \textbf{79.27} & 59.64 & -     & 86.17 & 44.64 & 77.71 & -     & \underline{68.73} & 37.11 & 20.92 & -     & \multicolumn{1}{c|}{-}                         & 80.96 \\
\multicolumn{1}{l|}{Show-o2}                                       & \multicolumn{1}{c|}{1.5B}                         & \underline{82.10} & \underline{78.31} & \textbf{97.28} & 96.78 & 97.68 & 98.25 & 40.83 & \underline{65.15} & \textbf{67.06} & \underline{94.81} & \underline{76.01} & \underline{95.20} & \underline{80.89} & 62.61 & \underline{57.67} & \textbf{23.29} & \textbf{25.27} & \multicolumn{1}{c|}{\underline{27.00}}                     & \underline{81.34} \\
\rowcolor[HTML]{ECF4FF} 
\multicolumn{1}{l|}{\cellcolor[HTML]{ECF4FF}\model} & \multicolumn{1}{c|}{\cellcolor[HTML]{ECF4FF}1.5B} & \textbf{84.32}     & \textbf{83.04}     & 95.99     & 96.72     & \textbf{98.02}     & \underline{98.33}     & \underline{69.39}     & \textbf{65.88}     & \underline{66.83}     & \textbf{95.41}     & \textbf{92.31}     & \textbf{97.50}     & \textbf{87.67}     & \textbf{78.12}     & \textbf{58.59}     &   \underline{23.18}   &  \underline{24.68}    & \multicolumn{1}{c|}{\cellcolor[HTML]{ECF4FF}\textbf{27.71}} & \textbf{84.06}     \\ \bottomrule
\end{tabular}
\vspace{-0.5em}
\caption{Video generation results on VBench. Full column names: QS: Quality Score, SS: Semantic Score, SC: Subject Consistency, BC: Background Consistency, TF: Temporal Flickering, MS: Motion Smoothness, DD: Dynamic Degree, AQ: Aesthetic Quality, IQ: Imaging Quality, OC: Object Class, MO: Multiple Objects, HA: Human Action, C: Color, SR: Spatial Relationship, S: Scene, AS: Appearance style, TS: Temporal Style, OC’: Overall Consistency. \textbf{Bold}: best results among each section. \underline{Underline}: second-best.}
\label{tab:vbench}
\vspace{-1em}
\end{table*}

\subsection{Ablation: Visual Representation Design}
\label{sec:ablation}

In this section, we conduct a series of ablation experiments to systematically assess the effectiveness of our model architecture and training pipeline. For all experiments, we use a lightweight variant of \model built on the Qwen2.5-1.5B LLM with a smaller flow matching head. Using this setup, we evaluate three visual representation designs:
\begin{itemize}
    \item [1.] Decoupled representations, using SigLIP 2 features for understanding and Wan 2.2 VAE latents for generation (denoted as ``Decoupled'').
    \item [2.] Show-o2-style unified representations, using a dual-path late fusion strategy to obtain the final representations (denoted as ``Show-o2''). A detailed explanation of this design can be found in Section~\ref{sec:discussion}.
    \item[3.] \model's unified representation, initialized from three different pretrained representation encoders: SigLIP \citep{zhai2023sigmoid}, SigLIP 2 \citep{tschannen2025siglip}, and DINOv3\footnote{Model code: \texttt{siglip-so400m-patch14-384}, \texttt{siglip2-so400m-patch16-512} and \texttt{dinov3-vith16plus-pretrain-lvd1689m}.}~\citep{simeoni2025dinov3}.
\end{itemize}
All models are trained on a subset of our training data using a two-stage training pipeline (corresponding to Stage 1 and Stage 3 in Section~\ref{sec:training}), with an equal number of training steps per stage. Our ablation study results are shown in Table~\ref{tab:ablation}. 

\noindent\textbf{Unified vs. decoupled visual representation.} Comparing Model 8 and Model 12 in Table~\ref{tab:ablation}, we observe that our unified representation consistently outperforms the decoupled setting across all understanding and generation benchmarks. Comparing Models 2 and 5 with Model 8, we find that training a unified model using decoupled visual representations results in substantial degradation on understanding tasks, compared to only training the model on understanding data. In contrast, Model 12 surpasses Model 3 on most understanding benchmarks and outperforms Model 6 across all generation benchmarks. These results indicate that our unified representation suffers far less from representation conflicts than decoupled designs, enabling stronger performance in both understanding and generation.

\noindent\textbf{Selection of representation encoders.} We find that \model's unified representation generally benefits from stronger representation encoders. As shown in Table~\ref{tab:ablation}, comparing Models 10, 12, and 13, both SigLIP 2 (400M parameters) and DINOv3 (800M) outperform SigLIP (400M) on all benchmarks. Furthermore, comparing Model 9 to Model 11 and Model 10 to Model 12, we observe that regardless of which VAE encoder is used in the model, replacing SigLIP with SigLIP 2 in the representation encoder consistently improves performance across all understanding and generation benchmarks. We ultimately adopt SigLIP 2 for \model as it delivers comparable understanding performance, superior generation quality relative to DINOv3, and maintains a significantly smaller model size.

\begin{table*}[!t]

\centering
\small

\setlength\tabcolsep{7.4 pt}
\begin{tabular}{l|c|c|ccccccp{1cm}@{}}
\toprule
\multirow{2}{*}{\textbf{Models}}          & \multirow{2}{*}{\textbf{ID}} & \multirow{2}{*}{\textbf{Data}}                                                & \multicolumn{4}{c}{\textbf{Understanding}}        & \multicolumn{2}{|c}{\textbf{Generation}} \\ \cmidrule(l){4-9} 
                                 &                     &                                                                      & \textbf{MME-p} & \textbf{MMMU} & \textbf{SEED} &\multicolumn{1}{c|}{\textbf{GQA}}  & \textbf{GenEval}         & \textbf{DPG}          \\ \midrule
Show-o2 (Wan 2.1 VAE + SigLIP)                         & 1                   &  & 1351     & 36.1   & 62.1    & \multicolumn{1}{c|}{56.8}          & -       & -      \\
Decoupled (SigLIP 2 only) & 2                   &                                                                      & 1392  &  38.2 & 62.9   & \multicolumn{1}{c|}{58.1}  & -               & -            \\
\model (Wan 2.2 VAE + SigLIP 2) & 3                   &       \multirow[t]{-2}{*}{\begin{tabular}[c]{@{}c@{}}Und.\\ Only\end{tabular}}                                                               & 1386  & 37.6 & 62.9   & \multicolumn{1}{c|}{57.4}  & -               & -            \\ \midrule
Show-o2 (Wan 2.1 VAE + SigLIP)                         & 4                   &  & -     & -   & -   & \multicolumn{1}{c|}{-}    & 76.2               & 82.56            \\
Decoupled (Wan 2.2 VAE only) & 5                   &                                                                     & -      & -  & -   & \multicolumn{1}{c|}{-}    & 77.3            & 82.87        \\
\model (Wan 2.2 VAE + SigLIP 2) & 6                   &    \multirow[t]{-2}{*}{\begin{tabular}[c]{@{}c@{}}Gen.\\ Only\end{tabular}}                                                                  & -      & -  & -   & \multicolumn{1}{c|}{-}    & 77.8            & 83.33        \\ \midrule
Show-o2 (Wan 2.1 VAE + SigLIP)                         & 7                   &                                                & 1339     & 35.4   & 61.7   & \multicolumn{1}{c|}{55.9}    & 75.9               & 82.32            \\
Decoupled (Wan 2.2 VAE + SigLIP 2)                        & 8                   &                                                                      & 1346  & 37.2  & 61.4 & \multicolumn{1}{c|}{56.5} & 78.3               & 83.50            \\
\model (Wan 2.1 VAE + SigLIP)   & 9                   &                                                                   & 1358  & 35.9 & 64.2 & \multicolumn{1}{c|}{57.2} & 77.2            & 83.29        \\
\model (Wan 2.2 VAE + SigLIP)   & 10                   &                                                                 & 1349  & 36.3 & 64.6 & \multicolumn{1}{c|}{57.4} & 76.9            & 83.10        \\
\model (Wan 2.1 VAE + SigLIP 2)   & 11                   &                                                                   & 1379  & 37.7 & 65.9  & \multicolumn{1}{c|}{58.4} & 79.1            & 83.98        \\
\rowcolor[HTML]{ECF4FF} 
\model (Wan 2.2 VAE + SigLIP 2) & 12                   &                                                                      & 1361  & 38.1  & 66.5  & \multicolumn{1}{c|}{58.2} & 79.4            & 84.20        \\
\model (Wan 2.2 VAE + DINOv3)   & 13                   &     \multirow[t]{-4}{*}{\begin{tabular}[c]{@{}c@{}}Und.\\\&\\Gen.\end{tabular}}                                                                 & 1396  & 37.3 & 65.6 & \multicolumn{1}{c|}{58.6} & 78.9            & 84.08        \\ \bottomrule
\end{tabular}

\vspace{-0.5em}
\caption{Ablation study results. ``Und. Only'', ``Gen. Only'' and ``Und. \& Gen.'' refer to models trained with understanding data only, generation data only, and both data, respectively.}
\label{tab:ablation}
\vspace{-1em}
\end{table*}

\noindent\textbf{Understanding-generation synergy.} Our experimental results on training models exclusively on either understanding (Models 1, 2 and 3) or generation data (Models 4, 5 and 6) demonstrate that \model benefits from joint training on both data types. Specifically, we observe that Model 12 surpasses Model 3 on understanding benchmarks and Model 6 on generation benchmarks. Although the comparison between Model 2 and Model 3 shows that \model's VAE + representation encoder architecture incurs a slight performance drop relative to using only a representation encoder (the standard setup for understanding-only models), our joint understanding + generation training pipeline largely compensates for this degradation. Specifically, Model 12 recovers its understanding performance and becomes comparable to or even better than Model 2 on several understanding benchmarks. Moreover, Model 12 substantially outperforms both Model 5 and Model 6 on all generation benchmarks. These results demonstrate the mutual enhancement between understanding and generation made possible by our unified visual representation design.

\noindent\textbf{Comparison with Show-o2.} A closely related work to ours is Show-o2~\citep{xie2025show}, which uses a dual-path late-fusion mechanism, merging features from separate VAE and semantic encoders via a fusion layer to build unified representations. In contrast, \model extracts unified representations directly from VAE latents using a semantic encoder, achieving deep feature fusion across all layers in the semantic encoder. Comparing Model 7 with Models 9, 10, 11 and 12 in Table~\ref{tab:ablation}, our unified representation consistently outperforms Show-o2 on all benchmarks, regardless of the choice of the VAE encoder and the representation encoder. Model 1 vs. 3 and Model 4 vs. 6 further show that Show-o2 underperforms even when trained on a single task. We attribute this to its late-fusion strategy, which introduces representation conflicts and degrades overall performance.

\subsection{Discussion: Unified Representation Analysis}
\label{sec:discussion}
As discussed in Section~\ref{sec:ablation}, both \model and Show-o2 \citep{xie2025show} employ unified visual representations for understanding and generation, but they construct these representations in fundamentally different ways. In this section, we first describe Show-o2's unified visual representation design in detail, and then provide an in-depth analysis of why \model's unified representation achieves superior performance than Show-o2.

\begin{figure}[t!]
\centering
\small
\includegraphics[width=0.7\linewidth]{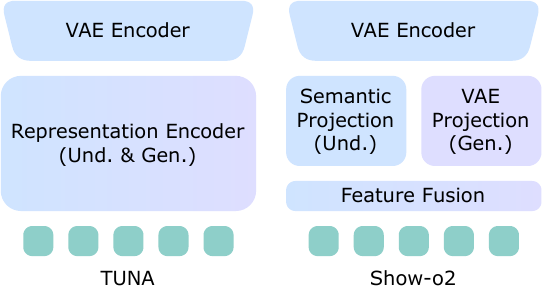}
\vspace{-1.5em}
\caption{Comparison between \model and Show-o2 on how unified visual representations are produced.}
\vspace{-1em}
\label{fig:arch_compare}
\end{figure}

As illustrated in Figure~\ref{fig:arch_compare}, Show-o2 constructs unified visual representations using a dual-path feature fusion mechanism. The input image or video is first encoded by a VAE encoder, after which the latent is processed through two parallel branches. The semantic projection branch feeds the VAE latents into a set of semantic layers to extract features for understanding tasks. The VAE projection branch applies 2D patch embedding layers to produce features tailored for generation tasks. Importantly, the semantic layers are \textit{pre-distilled} using a frozen representation encoder: given the same image, their outputs are first aligned with a pretrained SigLIP model before conducting end-to-end training of the Show-o2 model. This pre-distillation stage is proposed to preserve semantic understanding capability. Finally, Show-o2 merges the outputs of both paths using a feature fusion layer to obtain its unified visual representation.

\begin{figure}[t]
    \centering
    \begin{subfigure}[b]{0.35\linewidth}
        \includegraphics[width=\linewidth]{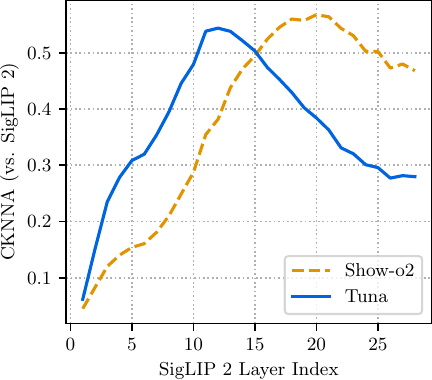}
        \caption{Alignment to SigLIP 2}
        \label{fig:cknna_siglip2}
    \end{subfigure}
    \hspace{2em}
    \begin{subfigure}[b]{0.35\linewidth}
        \includegraphics[width=\linewidth]{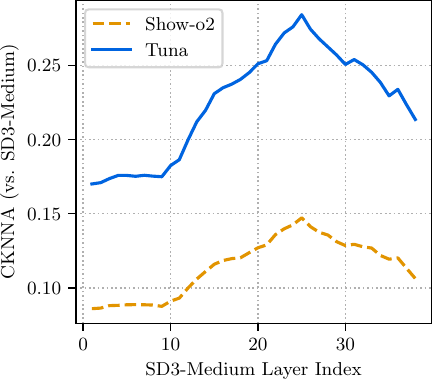}
        \caption{Alignment to SD3-Medium}
        \label{fig:cknna_sd3}
    \end{subfigure}
    \vspace{-0.8em}
    \caption{Representation alignment analysis with SigLIP 2 and SD3-Medium. For both \model and Show-o2, we extract visual representations at the input layer of the LLM decoder.}
    \vspace{-1em}
\end{figure}

To better understand why our unified representation yields superior performance, we perform a representation alignment analysis using CKNNA scores~\citep{huh2024platonic} with respect to two reference models: (1) a strong semantic encoder SigLIP 2~\citep{tschannen2025siglip}, and (2) a strong generation model SD3-Medium~\citep{esser2024scaling}. Concretely, we extract unified visual representations from \model and Show-o2 based on 1,024 images from the Wikipedia Captions dataset~\citep{srinivasan2021wit} and compute their CKNNA scores relative to features from all intermediate layers of the two reference models. The results are presented in Figure~\ref{fig:cknna_siglip2} and Figure~\ref{fig:cknna_sd3}.

As shown in the figures, both \model and Show-o2 exhibit strong alignment with the SigLIP 2 intermediate features, with CKNNA scores exceeding 0.5. This high similarity reflects their strong semantic understanding capability, consistent with their solid performance on multimodal understanding tasks. On the other hand, \model's unified representation shows consistently higher alignment with the SD3-Medium intermediate features compared to Show-o2, indicating that \model learns a more balanced unified representation suited for both understanding and generation. In contrast, Show-o2 remains biased toward semantic features, which limits its generation quality.

The above findings prompt us to further investigate why Show-o2's dual-path fusion mechanism produces biased features toward semantic understanding. To analyze this, we compute CKNNA scores between Show-o2's final fused features and the intermediate features from its understanding (semantic projection) and generation (VAE projection) branches before fusion. We find that Show-o2's unified representation exhibits a strong correlation with its understanding branch (CKNNA$=$0.45) but a very weak correlation with the generation branch (CKNNA$=$0.07). This demonstrates that the late-fusion strategy merges features in an imbalanced manner, causing the representation to remain dominated by semantic information. In contrast, \model's end-to-end training of the unified representation on both objectives enables early fusion of understanding and generation signals at every layer of the representation encoder. This layer-wise interaction captures richer cross-task dependencies and is inherently more robust than the late-fusion strategy adopted in Show-o2.

\begin{figure*}[t!]
\centering
\small
\includegraphics[width=\linewidth]{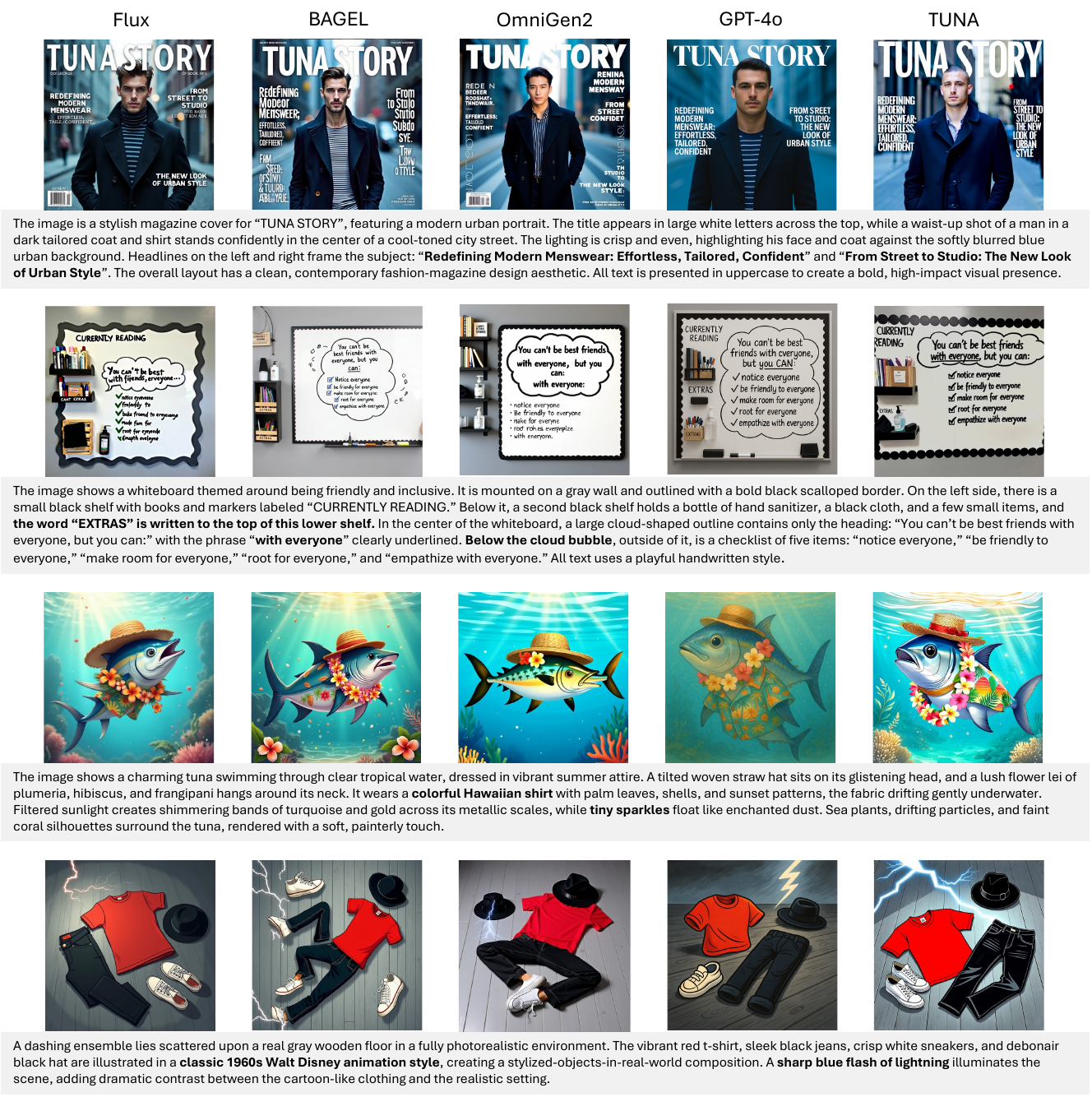}
\vspace{-2em}
\caption{Qualitative comparison between \model and baseline models on image generation tasks. The instructions that are correctly reflected in our results but failed in some of the baseline models are \textbf{bolded}.}
\vspace{-1em}
\label{fig:gen_compare}
\end{figure*}

\begin{figure*}[t!]
\centering
\small
\includegraphics[width=\linewidth]{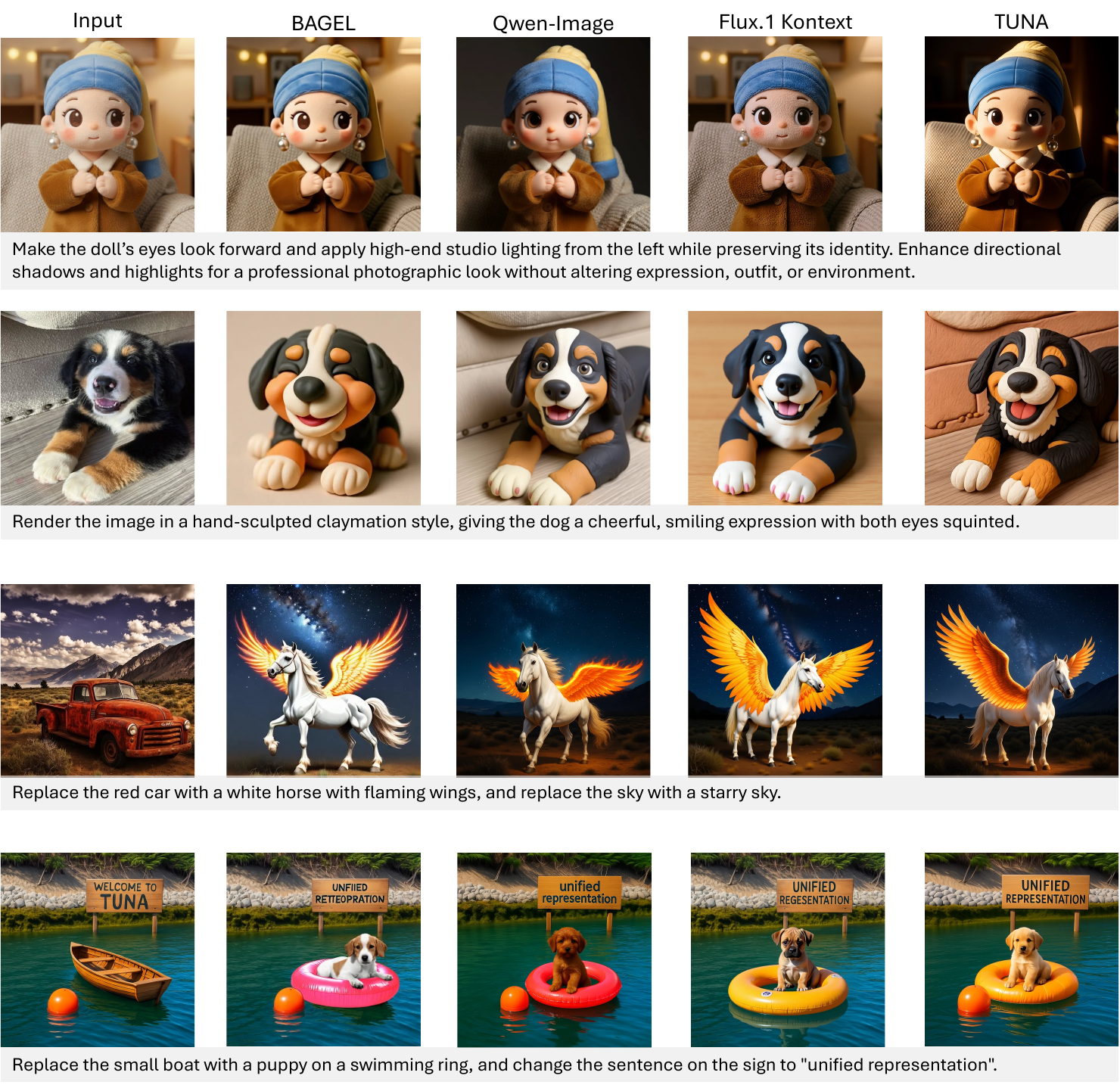}
\vspace{-2em}
\caption{Qualitative comparison between \model and baseline models on image editing tasks.}
\vspace{-1em}
\label{fig:edit_compare}
\end{figure*}

\begin{figure*}[t!]
\centering
\small
\includegraphics[width=\linewidth]{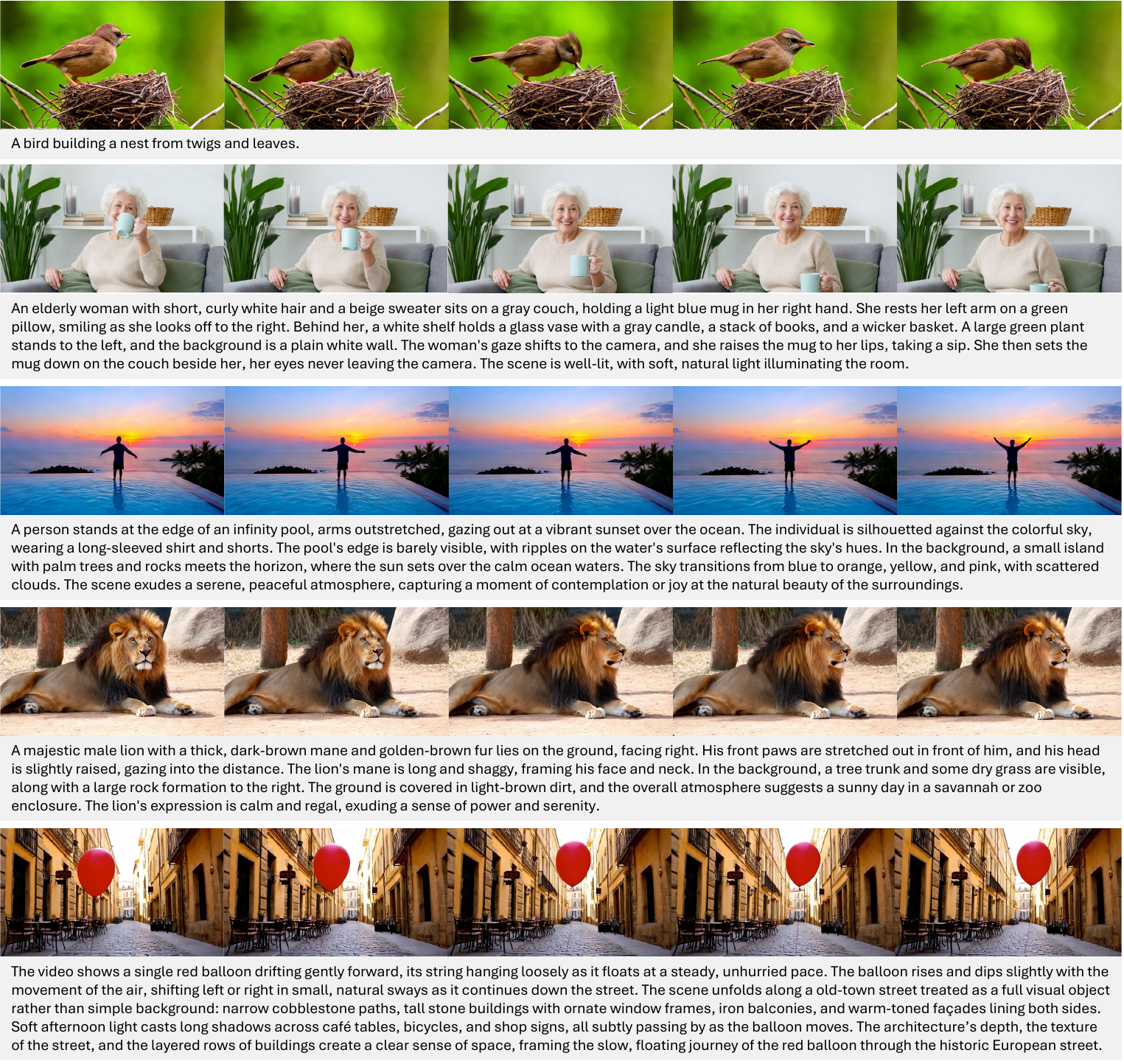}
\vspace{-2em}
\caption{Qualitative results for \model on the task of text-to-video generation.}
\vspace{-1em}
\label{fig:video_gen}
\end{figure*}

\subsection{Qualitative Results}
\noindent\textbf{Image generation.} We compare \model with state-of-the-art generation-only and unified models across diverse image generation instructions in Figure~\ref{fig:gen_compare}. In the first two examples, \model exhibits strong text rendering ability, accurately reproducing all visual text in the prompts without errors. In the whiteboard example, \model is the only model that correctly places an underline beneath ``with everyone'', demonstrating precise prompt-following capability. Moreover, \model accurately generates two black shelves, one containing books and markers on top and the other containing black cloth and hand sanitizer at the bottom, each in the correct position. Other models either fail to produce the correct number of shelves or place the wrong items on them. These results show that \model excels at compositional image generation, enabled by its unified visual representation with strong semantic understanding capabilities. In the “tuna” example, both \model and Flux~\citep{batifol2025flux} successfully render the Hawaiian shirt, while other models either fail to depict the shirt or generate an incorrect tuna body. Finally, in the ``red t-shirt'' example, \model accurately reflects the ``classic 1960s Walt Disney animation style'' and correctly includes all required elements from the prompt, maintaining a coherent and well-structured composition.

\noindent\textbf{Image editing.} We compare \model with BAGEL~\citep{deng2025emerging}, Qwen-Image~\citep{wu2025qwen}, and Flux.1 Kontext~\citep{batifol2025flux} on image editing tasks in Figure~\ref{fig:edit_compare}. As shown, \model not only correctly performs explicit editing operations, such as style transfer (photorealistic $\rightarrow$ hand-sculpted claymation in the ``dog'' example), environment change (daylight $\rightarrow$ nighttime in the ``red car'' example), and object replacement (boat $\rightarrow$ puppy with a swim ring in the ``boat'' example), but also handles more implicit and nuanced instructions, such as applying lighting from the left side in the ``doll'' example. These results further highlight \model's strong semantic understanding and high-fidelity image generation capabilities.

\noindent\textbf{Video generation.} We present \model's video generation results in Figure~\ref{fig:video_gen}. The model produces high-fidelity videos across a wide range of instructions, demonstrating the strength of its unified visual representation space for jointly modeling both images and videos.

\section{Related Work}
\label{sec:related}

\subsection{Large Multimodal Models}
Large multimodal models (LMMs) aim to generate text responses from multimodal inputs spanning images, videos, and text. Early LMMs such as Flamingo~\citep{alayrac2022flamingo} and Idefics~\citep{laurenccon2023obelics} introduced cross-attention layers to enable interaction between visual and linguistic features. Modern LMMs generally follow the LLaVA paradigm~\citep{liu2023visual}, where visual inputs are encoded by a vision encoder (e.g., CLIP~\citep{radford2021learning}) and then concatenated with text tokens for joint processing by a language model decoder. Recent research advances focus on improving instruction-following through higher-quality training data~\citep{llavanext2024, li2024llava, chen2024sharegpt4v, li2024mvbench, ren2024vista, zhang2024video, wiedmann2025finevision, an2025llava}, developing stronger vision encoders capable of handling higher-resolution images~\citep{llavanext2024, laurenccon2024matters, wang2024qwen2, bai2025qwen2}, extending LMMs to interleaved image~\citep{laurenccon2024matters, li2024llava, jiang2024mantis} and video understanding~\citep{maaz2023video, lin2023video, zhang2024long, li2024mvbench, li2024videochat, ren2025vamba}, and incorporating reinforcement learning with thinking modes~\citep{deng2025openvlthinker, huang2025vision, feng2025video} or pixel-space reasoning~\citep{su2025pixel, su2025thinking, liu2025visionreasoner}.

\subsection{Diffusion Generative Models}
Diffusion generative models have become the de facto backbone of high-fidelity image \citep{esser2024scaling, batifol2025flux, li2024hunyuan, wu2025qwen,liu2023cones,liu2023customizable,liu2025manganinja} and video \citep{kong2024hunyuanvideo, seawead2025seaweed, wan2025wan, liu2025worldweaver} synthesis. Modern large-scale visual generation models typically apply diffusion in a continuous latent space defined by a learned VAE, following the Latent Diffusion Model (LDM) paradigm~\citep{rombach2022high}, which offers superior perceptual quality and sampling efficiency compared to autoregressive decoding of long sequences of discrete tokens based on VQ-VAE \citep{van2017neural, esser2021taming}. Within diffusion itself, latent-space models \citep{rombach2022high, podell2023sdxl, esser2024scaling} are generally preferred over pixel-space approaches~\citep{dhariwal2021diffusion, saharia2022photorealistic} because they reduce computational cost, ease scaling to higher resolutions, and allow the denoising network to focus on semantically meaningful structure rather than low-level pixel noise. Architecturally, diffusion backbones have evolved from convolutional U-Net designs~\citep{ronneberger2015u, ho2020denoising} to diffusion transformers (DiT) \citep{peebles2023scalable, ma2024sit}; In parallel, the learning objective has been generalized from Gaussian noise prediction and score matching~\citep{ho2020denoising, song2020score} to more expressive formulations such as rectified flows \citep{liu2022flow} and flow matching objectives~\citep{lipman2022flow, albergo2023stochastic}.

\subsection{Unified Multimodal Models}
Unified multimodal models (UMMs) have gained growing attention for their ability to flexibly generate text and visual content from diverse multimodal inputs. Recent approaches~\citep{luo2025next} such as MetaQuery \citep{pan2025transfer}, BLIP-3o \citep{chen2025blip3}, and UniWorld-V1 \citep{lin2025uniworld} achieve this by connecting understanding-only and generation-only models through learnable adapters. While achieving promising results, their capabilities largely rely on pretrained task-specific models, limiting the potential synergy between understanding and generation. In contrast, native UMMs are pretrained from scratch to perform both tasks within a single unified architecture. Among these works, models such as the Janus series \citep{wu2025janus, ma2025janusflow, chen2025janus} and UniFluid \citep{fan2025unified} adopt decoupled visual representations for understanding and generation. BAGEL \citep{deng2025emerging}, Mogao \citep{liao2025mogao}, and OneCAT \citep{li2025onecat} further use MoE-style architectures to route inputs separately, mitigating conflicts between distinct representations from the decoupled vision encoders. Alternatively, models like Chameleon \citep{team2024chameleon}, Transfusion \citep{zhou2024transfusion}, Harmon \citep{wu2025harmonizing}, and the Show-o series \citep{xie2024show, xie2025show} employ unified visual representations for both tasks. While being more efficient, these models often exhibit weaker or imbalanced performance, excelling in one task but underperforming in the other. \model overcomes these limitations by learning balanced, unified visual representations and achieves strong performance in both understanding and generation.

\subsection{Representation in Multimodal Models}
Recent studies have explored learning better representations to enhance multimodal understanding and generation models. From the perspective of improving understanding models, methods such as Ross \citep{wang2024reconstructive}, GenHancer \citep{ma2025genhancer} and ASVR \citep{wang2025autoregressive} enhance multimodal understanding by introducing generation or reconstruction objectives, encouraging the model to capture fine-grained visual details. Conversely, to improve generative models, approaches such as REPA \citep{yu2024representation} and VA-VAE \citep{yao2025reconstruction} align diffusion transformers or VAE representations with semantic vision encoders, thereby achieving stronger generative performance. Similarly, Dispersive Loss \citep{wang2025diffuse} introduces an auxiliary contrastive-like objective to further enhance generation quality.

In the domain of unified multimodal models, recent research has primarily focused on developing unified visual tokenizers that support both understanding and generation tasks. For instance, TokenFlow~\citep{qu2025tokenflow} and MUSE-VL~\citep{xie2024muse} adopt late-fusion strategies to merge features from separate understanding and generation encoders into quantized codebooks. DualToken~\citep{song2025dualtoken}, UniTok~\citep{ma2025unitok} and TokLIP~\citep{lin2025toklipmarryvisualtokens} train a single encoder to produce vector-quantized representations for both tasks. However, these methods rely on discrete representations, limiting their ability to perform high-fidelity visual generation. 
UniFlow~\citep{yue2025uniflow} and UniLIP~\citep{tang2025unilip} adapt representation encoders into continuous unified visual tokenizers, but both rely on relatively complex alignment (\eg self-distillation or reconstruction schemes). In contrast, TUNA learns a unified representation end-to-end under joint understanding and generation objectives, and is validated at a larger scale across more tasks. Moreover, UniLIP adopts a composite design where the unified features only serve as conditions for a separate pretrained generative model (SANA~\citep{xie2024sana}). On the other hand, \model trains a native unified model that jointly performs understanding and generation within a single framework.
\section{Conclusion}
\label{sec:conclusion}
We introduced \model, a native unified multimodal model that constructs a unified visual representation space by cascading a VAE encoder with a representation encoder. We train an LLM decoder and a flow matching head on this unified representation, achieving strong performance across image and video understanding, image and video generation, and image editing. \model not only surpasses prior UMM baselines but also performs competitively with leading understanding-only and generation-only models. Our ablation studies further show that (1) \model's unified representation space outperforms both Show-o2-style unified representations and decoupled representation designs, (2) stronger pretrained representation encoders consistently yield better performance within our framework, and (3) our unified visual representation design enables mutual enhancement between understanding and generation.

\section{Acknowledgment}
We would like to thank Yukang Yang (Princeton University), Ji Xie (UC Berkeley), and Jinheng Xie (NUS) for their constructive feedback on this project.
\clearpage
\newpage
\bibliographystyle{assets/plainnat}
\bibliography{paper}

@article{zhou2024transfusion,
  title={Transfusion: Predict the next token and diffuse images with one multi-modal model},
  author={Zhou, Chunting and Yu, Lili and Babu, Arun and Tirumala, Kushal and Yasunaga, Michihiro and Shamis, Leonid and Kahn, Jacob and Ma, Xuezhe and Zettlemoyer, Luke and Levy, Omer},
  journal={arXiv preprint arXiv:2408.11039},
  year={2024}
}

@article{deng2025emerging,
  title={Emerging properties in unified multimodal pretraining},
  author={Deng, Chaorui and Zhu, Deyao and Li, Kunchang and Gou, Chenhui and Li, Feng and Wang, Zeyu and Zhong, Shu and Yu, Weihao and Nie, Xiaonan and Song, Ziang and others},
  journal={arXiv preprint arXiv:2505.14683},
  year={2025}
}

@article{xie2025show,
  title={Show-o2: Improved Native Unified Multimodal Models},
  author={Xie, Jinheng and Yang, Zhenheng and Shou, Mike Zheng},
  journal={arXiv preprint arXiv:2506.15564},
  year={2025}
}

@article{wang2024reconstructive,
  title={Reconstructive visual instruction tuning},
  author={Wang, Haochen and Zheng, Anlin and Zhao, Yucheng and Wang, Tiancai and Ge, Zheng and Zhang, Xiangyu and Zhang, Zhaoxiang},
  journal={arXiv preprint arXiv:2410.09575},
  year={2024}
}

@article{yu2024representation,
  title={Representation alignment for generation: Training diffusion transformers is easier than you think},
  author={Yu, Sihyun and Kwak, Sangkyung and Jang, Huiwon and Jeong, Jongheon and Huang, Jonathan and Shin, Jinwoo and Xie, Saining},
  journal={arXiv preprint arXiv:2410.06940},
  year={2024}
}

@article{oquab2023dinov2,
  title={Dinov2: Learning robust visual features without supervision},
  author={Oquab, Maxime and Darcet, Timoth{\'e}e and Moutakanni, Th{\'e}o and Vo, Huy and Szafraniec, Marc and Khalidov, Vasil and Fernandez, Pierre and Haziza, Daniel and Massa, Francisco and El-Nouby, Alaaeldin and others},
  journal={arXiv preprint arXiv:2304.07193},
  year={2023}
}

@article{chen2025janus,
  title={Janus-pro: Unified multimodal understanding and generation with data and model scaling},
  author={Chen, Xiaokang and Wu, Zhiyu and Liu, Xingchao and Pan, Zizheng and Liu, Wen and Xie, Zhenda and Yu, Xingkai and Ruan, Chong},
  journal={arXiv preprint arXiv:2501.17811},
  year={2025}
}

@inproceedings{radford2021learning,
  title={Learning transferable visual models from natural language supervision},
  author={Radford, Alec and Kim, Jong Wook and Hallacy, Chris and Ramesh, Aditya and Goh, Gabriel and Agarwal, Sandhini and Sastry, Girish and Askell, Amanda and Mishkin, Pamela and Clark, Jack and others},
  booktitle={International conference on machine learning},
  pages={8748--8763},
  year={2021},
  organization={PmLR}
}

@article{tschannen2025siglip,
  title={Siglip 2: Multilingual vision-language encoders with improved semantic understanding, localization, and dense features},
  author={Tschannen, Michael and Gritsenko, Alexey and Wang, Xiao and Naeem, Muhammad Ferjad and Alabdulmohsin, Ibrahim and Parthasarathy, Nikhil and Evans, Talfan and Beyer, Lucas and Xia, Ye and Mustafa, Basil and others},
  journal={arXiv preprint arXiv:2502.14786},
  year={2025}
}

@inproceedings{rombach2022high,
  title={High-resolution image synthesis with latent diffusion models},
  author={Rombach, Robin and Blattmann, Andreas and Lorenz, Dominik and Esser, Patrick and Ommer, Bj{\"o}rn},
  booktitle={Proceedings of the IEEE/CVF conference on computer vision and pattern recognition},
  pages={10684--10695},
  year={2022}
}

@inproceedings{esser2021taming,
  title={Taming transformers for high-resolution image synthesis},
  author={Esser, Patrick and Rombach, Robin and Ommer, Bjorn},
  booktitle={Proceedings of the IEEE/CVF conference on computer vision and pattern recognition},
  pages={12873--12883},
  year={2021}
}

@inproceedings{zhai2023sigmoid,
  title={Sigmoid loss for language image pre-training},
  author={Zhai, Xiaohua and Mustafa, Basil and Kolesnikov, Alexander and Beyer, Lucas},
  booktitle={Proceedings of the IEEE/CVF international conference on computer vision},
  pages={11975--11986},
  year={2023}
}

@article{wan2025wan,
  title={Wan: Open and advanced large-scale video generative models},
  author={Wan, Team and Wang, Ang and Ai, Baole and Wen, Bin and Mao, Chaojie and Xie, Chen-Wei and Chen, Di and Yu, Feiwu and Zhao, Haiming and Yang, Jianxiao and others},
  journal={arXiv preprint arXiv:2503.20314},
  year={2025}
}

@article{pan2025transfer,
  title={Transfer between modalities with metaqueries},
  author={Pan, Xichen and Shukla, Satya Narayan and Singh, Aashu and Zhao, Zhuokai and Mishra, Shlok Kumar and Wang, Jialiang and Xu, Zhiyang and Chen, Jiuhai and Li, Kunpeng and Juefei-Xu, Felix and others},
  journal={arXiv preprint arXiv:2504.06256},
  year={2025}
}

@article{chen2025blip3,
  title={Blip3-o: A family of fully open unified multimodal models-architecture, training and dataset},
  author={Chen, Jiuhai and Xu, Zhiyang and Pan, Xichen and Hu, Yushi and Qin, Can and Goldstein, Tom and Huang, Lifu and Zhou, Tianyi and Xie, Saining and Savarese, Silvio and others},
  journal={arXiv preprint arXiv:2505.09568},
  year={2025}
}

@article{lin2025uniworld,
  title={Uniworld: High-resolution semantic encoders for unified visual understanding and generation},
  author={Lin, Bin and Li, Zongjian and Cheng, Xinhua and Niu, Yuwei and Ye, Yang and He, Xianyi and Yuan, Shenghai and Yu, Wangbo and Wang, Shaodong and Ge, Yunyang and others},
  journal={arXiv preprint arXiv:2506.03147},
  year={2025}
}

@inproceedings{wu2025janus,
  title={Janus: Decoupling visual encoding for unified multimodal understanding and generation},
  author={Wu, Chengyue and Chen, Xiaokang and Wu, Zhiyu and Ma, Yiyang and Liu, Xingchao and Pan, Zizheng and Liu, Wen and Xie, Zhenda and Yu, Xingkai and Ruan, Chong and others},
  booktitle={Proceedings of the Computer Vision and Pattern Recognition Conference},
  pages={12966--12977},
  year={2025}
}

@inproceedings{ma2025janusflow,
  title={Janusflow: Harmonizing autoregression and rectified flow for unified multimodal understanding and generation},
  author={Ma, Yiyang and Liu, Xingchao and Chen, Xiaokang and Liu, Wen and Wu, Chengyue and Wu, Zhiyu and Pan, Zizheng and Xie, Zhenda and Zhang, Haowei and Yu, Xingkai and others},
  booktitle={Proceedings of the Computer Vision and Pattern Recognition Conference},
  pages={7739--7751},
  year={2025}
}

@article{xie2024muse,
  title={Muse-vl: Modeling unified vlm through semantic discrete encoding},
  author={Xie, Rongchang and Du, Chen and Song, Ping and Liu, Chang},
  journal={arXiv preprint arXiv:2411.17762},
  year={2024}
}

@article{liao2025mogao,
  title={Mogao: An omni foundation model for interleaved multi-modal generation},
  author={Liao, Chao and Liu, Liyang and Wang, Xun and Luo, Zhengxiong and Zhang, Xinyu and Zhao, Wenliang and Wu, Jie and Li, Liang and Tian, Zhi and Huang, Weilin},
  journal={arXiv preprint arXiv:2505.05472},
  year={2025}
}

@article{li2025onecat,
  title={Onecat: Decoder-only auto-regressive model for unified understanding and generation},
  author={Li, Han and Peng, Xinyu and Wang, Yaoming and Peng, Zelin and Chen, Xin and Weng, Rongxiang and Wang, Jingang and Cai, Xunliang and Dai, Wenrui and Xiong, Hongkai},
  journal={arXiv preprint arXiv:2509.03498},
  year={2025}
}

@article{team2024chameleon,
  title={Chameleon: Mixed-modal early-fusion foundation models},
  author={Team, Chameleon},
  journal={arXiv preprint arXiv:2405.09818},
  year={2024}
}

@article{wu2025harmonizing,
  title={Harmonizing visual representations for unified multimodal understanding and generation},
  author={Wu, Size and Zhang, Wenwei and Xu, Lumin and Jin, Sheng and Wu, Zhonghua and Tao, Qingyi and Liu, Wentao and Li, Wei and Loy, Chen Change},
  journal={arXiv preprint arXiv:2503.21979},
  year={2025}
}

@article{xie2024show,
  title={Show-o: One single transformer to unify multimodal understanding and generation},
  author={Xie, Jinheng and Mao, Weijia and Bai, Zechen and Zhang, David Junhao and Wang, Weihao and Lin, Kevin Qinghong and Gu, Yuchao and Chen, Zhijie and Yang, Zhenheng and Shou, Mike Zheng},
  journal={arXiv preprint arXiv:2408.12528},
  year={2024}
}

@article{grattafiori2024llama,
  title={The llama 3 herd of models},
  author={Grattafiori, Aaron and Dubey, Abhimanyu and Jauhri, Abhinav and Pandey, Abhinav and Kadian, Abhishek and Al-Dahle, Ahmad and Letman, Aiesha and Mathur, Akhil and Schelten, Alan and Vaughan, Alex and others},
  journal={arXiv preprint arXiv:2407.21783},
  year={2024}
}

@article{yang2025qwen3,
  title={Qwen3 technical report},
  author={Yang, An and Li, Anfeng and Yang, Baosong and Zhang, Beichen and Hui, Binyuan and Zheng, Bo and Yu, Bowen and Gao, Chang and Huang, Chengen and Lv, Chenxu and others},
  journal={arXiv preprint arXiv:2505.09388},
  year={2025}
}

@article{sun2024autoregressive,
  title={Autoregressive model beats diffusion: Llama for scalable image generation},
  author={Sun, Peize and Jiang, Yi and Chen, Shoufa and Zhang, Shilong and Peng, Bingyue and Luo, Ping and Yuan, Zehuan},
  journal={arXiv preprint arXiv:2406.06525},
  year={2024}
}

@article{nie2025large,
  title={Large language diffusion models},
  author={Nie, Shen and Zhu, Fengqi and You, Zebin and Zhang, Xiaolu and Ou, Jingyang and Hu, Jun and Zhou, Jun and Lin, Yankai and Wen, Ji-Rong and Li, Chongxuan},
  journal={arXiv preprint arXiv:2502.09992},
  year={2025}
}

@article{batifol2025flux,
  title={FLUX. 1 Kontext: Flow Matching for In-Context Image Generation and Editing in Latent Space},
  author={Batifol, Stephen and Blattmann, Andreas and Boesel, Frederic and Consul, Saksham and Diagne, Cyril and Dockhorn, Tim and English, Jack and English, Zion and Esser, Patrick and Kulal, Sumith and others},
  journal={arXiv e-prints},
  pages={arXiv--2506},
  year={2025}
}

@article{bai2025qwen2,
  title={Qwen2. 5-vl technical report},
  author={Bai, Shuai and Chen, Keqin and Liu, Xuejing and Wang, Jialin and Ge, Wenbin and Song, Sibo and Dang, Kai and Wang, Peng and Wang, Shijie and Tang, Jun and others},
  journal={arXiv preprint arXiv:2502.13923},
  year={2025}
}

@article{wang2025internvl3,
  title={Internvl3. 5: Advancing open-source multimodal models in versatility, reasoning, and efficiency},
  author={Wang, Weiyun and Gao, Zhangwei and Gu, Lixin and Pu, Hengjun and Cui, Long and Wei, Xingguang and Liu, Zhaoyang and Jing, Linglin and Ye, Shenglong and Shao, Jie and others},
  journal={arXiv preprint arXiv:2508.18265},
  year={2025}
}

@inproceedings{esser2024scaling,
  title={Scaling rectified flow transformers for high-resolution image synthesis},
  author={Esser, Patrick and Kulal, Sumith and Blattmann, Andreas and Entezari, Rahim and M{\"u}ller, Jonas and Saini, Harry and Levi, Yam and Lorenz, Dominik and Sauer, Axel and Boesel, Frederic and others},
  booktitle={Forty-first international conference on machine learning},
  year={2024}
}

@article{li2024autoregressive,
  title={Autoregressive image generation without vector quantization},
  author={Li, Tianhong and Tian, Yonglong and Li, He and Deng, Mingyang and He, Kaiming},
  journal={Advances in Neural Information Processing Systems},
  volume={37},
  pages={56424--56445},
  year={2024}
}

@article{zheng2025diffusion,
  title={Diffusion Transformers with Representation Autoencoders},
  author={Zheng, Boyang and Ma, Nanye and Tong, Shengbang and Xie, Saining},
  journal={arXiv preprint arXiv:2510.11690},
  year={2025}
}

@article{ma2025unitok,
  title={Unitok: A unified tokenizer for visual generation and understanding},
  author={Ma, Chuofan and Jiang, Yi and Wu, Junfeng and Yang, Jihan and Yu, Xin and Yuan, Zehuan and Peng, Bingyue and Qi, Xiaojuan},
  journal={arXiv preprint arXiv:2502.20321},
  year={2025}
}

@misc{lin2025toklipmarryvisualtokens,
      title={TokLIP: Marry Visual Tokens to CLIP for Multimodal Comprehension and Generation}, 
      author={Haokun Lin and Teng Wang and Yixiao Ge and Yuying Ge and Zhichao Lu and Ying Wei and Qingfu Zhang and Zhenan Sun and Ying Shan},
      year={2025},
      eprint={2505.05422},
      archivePrefix={arXiv},
      primaryClass={cs.CV},
      url={https://arxiv.org/abs/2505.05422}, 
}

@inproceedings{li2023your,
  title={Your diffusion model is secretly a zero-shot classifier},
  author={Li, Alexander C and Prabhudesai, Mihir and Duggal, Shivam and Brown, Ellis and Pathak, Deepak},
  booktitle={Proceedings of the IEEE/CVF International Conference on Computer Vision},
  pages={2206--2217},
  year={2023}
}

@article{zuo2024scenediff,
  title={SceneDiff: Generative scene-level image retrieval with text and sketch using diffusion models},
  author={Zuo, Ran and Hu, Haoxiang and Deng, Xiaoming and Gao, Cangjun and Zhang, Zhengming and Lai, Yukun and Ma, Cuixia and Liu, Yong-Jin and Wang, Hongan},
  year={2024}
}

@article{zhu2024unleashing,
  title={Unleashing the potential of the diffusion model in few-shot semantic segmentation},
  author={Zhu, Muzhi and Liu, Yang and Luo, Zekai and Jing, Chenchen and Chen, Hao and Xu, Guangkai and Wang, Xinlong and Shen, Chunhua},
  journal={Advances in Neural Information Processing Systems},
  volume={37},
  pages={42672--42695},
  year={2024}
}

@software{Wan2.2,
  author       = {Wan-Video team},
  title        = {Wan: Open and Advanced Large-Scale Video Generative Models (Wan2.2)},
  year         = {2025},
  url          = {https://github.com/Wan-Video/Wan2.2},
  note         = {GitHub repository, Apache-2.0 License}
}

@inproceedings{peebles2023scalable,
  title={Scalable diffusion models with transformers},
  author={Peebles, William and Xie, Saining},
  booktitle={Proceedings of the IEEE/CVF international conference on computer vision},
  pages={4195--4205},
  year={2023}
}

@article{tschannen2023image,
  title={Image captioners are scalable vision learners too},
  author={Tschannen, Michael and Kumar, Manoj and Steiner, Andreas and Zhai, Xiaohua and Houlsby, Neil and Beyer, Lucas},
  journal={Advances in Neural Information Processing Systems},
  volume={36},
  pages={46830--46855},
  year={2023}
}

@article{fan2025unified,
  title={Unified autoregressive visual generation and understanding with continuous tokens},
  author={Fan, Lijie and Tang, Luming and Qin, Siyang and Li, Tianhong and Yang, Xuan and Qiao, Siyuan and Steiner, Andreas and Sun, Chen and Li, Yuanzhen and Zhu, Tao and others},
  journal={arXiv preprint arXiv:2503.13436},
  year={2025}
}

@article{ma2025genhancer,
  title={GenHancer: Imperfect generative models are secretly strong vision-centric enhancers},
  author={Ma, Shijie and Ge, Yuying and Wang, Teng and Guo, Yuxin and Ge, Yixiao and Shan, Ying},
  journal={arXiv preprint arXiv:2503.19480},
  year={2025}
}

@article{wang2025autoregressive,
  title={Autoregressive semantic visual reconstruction helps vlms understand better},
  author={Wang, Dianyi and Song, Wei and Wang, Yikun and Wang, Siyuan and Yu, Kaicheng and Wei, Zhongyu and Wang, Jiaqi},
  journal={arXiv preprint arXiv:2506.09040},
  year={2025}
}

@inproceedings{yao2025reconstruction,
  title={Reconstruction vs. generation: Taming optimization dilemma in latent diffusion models},
  author={Yao, Jingfeng and Yang, Bin and Wang, Xinggang},
  booktitle={Proceedings of the Computer Vision and Pattern Recognition Conference},
  pages={15703--15712},
  year={2025}
}

@article{wang2025diffuse,
  title={Diffuse and Disperse: Image Generation with Representation Regularization},
  author={Wang, Runqian and He, Kaiming},
  journal={arXiv preprint arXiv:2506.09027},
  year={2025}
}

@inproceedings{qu2025tokenflow,
  title={Tokenflow: Unified image tokenizer for multimodal understanding and generation},
  author={Qu, Liao and Zhang, Huichao and Liu, Yiheng and Wang, Xu and Jiang, Yi and Gao, Yiming and Ye, Hu and Du, Daniel K and Yuan, Zehuan and Wu, Xinglong},
  booktitle={Proceedings of the Computer Vision and Pattern Recognition Conference},
  pages={2545--2555},
  year={2025}
}

@article{song2025dualtoken,
  title={Dualtoken: Towards unifying visual understanding and generation with dual visual vocabularies},
  author={Song, Wei and Wang, Yuran and Song, Zijia and Li, Yadong and Sun, Haoze and Chen, Weipeng and Zhou, Zenan and Xu, Jianhua and Wang, Jiaqi and Yu, Kaicheng},
  journal={arXiv preprint arXiv:2503.14324},
  year={2025}
}

@article{xie2024sana,
  title={Sana: Efficient high-resolution image synthesis with linear diffusion transformers},
  author={Xie, Enze and Chen, Junsong and Chen, Junyu and Cai, Han and Tang, Haotian and Lin, Yujun and Zhang, Zhekai and Li, Muyang and Zhu, Ligeng and Lu, Yao and others},
  journal={arXiv preprint arXiv:2410.10629},
  year={2024}
}

@article{tang2025unilip,
  title={Unilip: Adapting clip for unified multimodal understanding, generation and editing},
  author={Tang, Hao and Xie, Chenwei and Bao, Xiaoyi and Weng, Tingyu and Li, Pandeng and Zheng, Yun and Wang, Liwei},
  journal={arXiv preprint arXiv:2507.23278},
  year={2025}
}

@article{yue2025uniflow,
  title={UniFlow: A Unified Pixel Flow Tokenizer for Visual Understanding and Generation},
  author={Yue, Zhengrong and Zhang, Haiyu and Zeng, Xiangyu and Chen, Boyu and Wang, Chenting and Zhuang, Shaobin and Dong, Lu and Du, KunPeng and Wang, Yi and Wang, Limin and others},
  journal={arXiv preprint arXiv:2510.10575},
  year={2025}
}

@article{wiedmann2025finevision,
  title={Finevision: Open data is all you need},
  author={Wiedmann, Luis and Zohar, Orr and Mahla, Amir and Wang, Xiaohan and Li, Rui and Frere, Thibaud and von Werra, Leandro and Gosthipaty, Aritra Roy and Marafioti, Andr{\'e}s},
  journal={arXiv preprint arXiv:2510.17269},
  year={2025}
}

@article{zhang2024video,
  title={Video instruction tuning with synthetic data},
  author={Zhang, Yuanhan and Wu, Jinming and Li, Wei and Li, Bo and Ma, Zejun and Liu, Ziwei and Li, Chunyuan},
  journal={arXiv preprint arXiv:2410.02713},
  year={2024}
}

@article{li2024videochat,
  title={Videochat-flash: Hierarchical compression for long-context video modeling},
  author={Li, Xinhao and Wang, Yi and Yu, Jiashuo and Zeng, Xiangyu and Zhu, Yuhan and Huang, Haian and Gao, Jianfei and Li, Kunchang and He, Yinan and Wang, Chenting and others},
  journal={arXiv preprint arXiv:2501.00574},
  year={2024}
}

@article{huh2024platonic,
  title={The platonic representation hypothesis},
  author={Huh, Minyoung and Cheung, Brian and Wang, Tongzhou and Isola, Phillip},
  journal={arXiv preprint arXiv:2405.07987},
  year={2024}
}

@inproceedings{srinivasan2021wit,
  title={Wit: Wikipedia-based image text dataset for multimodal multilingual machine learning},
  author={Srinivasan, Krishna and Raman, Karthik and Chen, Jiecao and Bendersky, Michael and Najork, Marc},
  booktitle={Proceedings of the 44th international ACM SIGIR conference on research and development in information retrieval},
  pages={2443--2449},
  year={2021}
}

@article{simeoni2025dinov3,
  title={Dinov3},
  author={Sim{\'e}oni, Oriane and Vo, Huy V and Seitzer, Maximilian and Baldassarre, Federico and Oquab, Maxime and Jose, Cijo and Khalidov, Vasil and Szafraniec, Marc and Yi, Seungeun and Ramamonjisoa, Micha{\"e}l and others},
  journal={arXiv preprint arXiv:2508.10104},
  year={2025}
}

@inproceedings{li2025dual,
  title={Dual diffusion for unified image generation and understanding},
  author={Li, Zijie and Li, Henry and Shi, Yichun and Farimani, Amir Barati and Kluger, Yuval and Yang, Linjie and Wang, Peng},
  booktitle={Proceedings of the Computer Vision and Pattern Recognition Conference},
  pages={2779--2790},
  year={2025}
}

@inproceedings{xie2025muse,
  title={Muse-vl: Modeling unified vlm through semantic discrete encoding},
  author={Xie, Rongchang and Du, Chen and Song, Ping and Liu, Chang},
  booktitle={Proceedings of the IEEE/CVF International Conference on Computer Vision},
  pages={24135--24146},
  year={2025}
}

@article{wang2024emu3,
  title={Emu3: Next-token prediction is all you need},
  author={Wang, Xinlong and Zhang, Xiaosong and Luo, Zhengxiong and Sun, Quan and Cui, Yufeng and Wang, Jinsheng and Zhang, Fan and Wang, Yueze and Li, Zhen and Yu, Qiying and others},
  journal={arXiv preprint arXiv:2409.18869},
  year={2024}
}

@article{liu2023visual,
  title={Visual instruction tuning},
  author={Liu, Haotian and Li, Chunyuan and Wu, Qingyang and Lee, Yong Jae},
  journal={Advances in neural information processing systems},
  volume={36},
  pages={34892--34916},
  year={2023}
}

@article{bai2023qwen,
  title={Qwen technical report},
  author={Bai, Jinze and Bai, Shuai and Chu, Yunfei and Cui, Zeyu and Dang, Kai and Deng, Xiaodong and Fan, Yang and Ge, Wenbin and Han, Yu and Huang, Fei and others},
  journal={arXiv preprint arXiv:2309.16609},
  year={2023}
}

@article{li2024llava,
  title={Llava-onevision: Easy visual task transfer},
  author={Li, Bo and Zhang, Yuanhan and Guo, Dong and Zhang, Renrui and Li, Feng and Zhang, Hao and Zhang, Kaichen and Zhang, Peiyuan and Li, Yanwei and Liu, Ziwei and others},
  journal={arXiv preprint arXiv:2408.03326},
  year={2024}
}

@inproceedings{li2025synergen,
  title={Synergen-vl: Towards synergistic image understanding and generation with vision experts and token folding},
  author={Li, Hao and Tian, Changyao and Shao, Jie and Zhu, Xizhou and Wang, Zhaokai and Zhu, Jinguo and Dou, Wenhan and Wang, Xiaogang and Li, Hongsheng and Lu, Lewei and others},
  booktitle={Proceedings of the Computer Vision and Pattern Recognition Conference},
  pages={29767--29779},
  year={2025}
}

@article{wu2024vila,
  title={Vila-u: a unified foundation model integrating visual understanding and generation},
  author={Wu, Yecheng and Zhang, Zhuoyang and Chen, Junyu and Tang, Haotian and Li, Dacheng and Fang, Yunhao and Zhu, Ligeng and Xie, Enze and Yin, Hongxu and Yi, Li and others},
  journal={arXiv preprint arXiv:2409.04429},
  year={2024}
}

@misc{fu2025mmecomprehensiveevaluationbenchmark,
      title={MME: A Comprehensive Evaluation Benchmark for Multimodal Large Language Models}, 
      author={Chaoyou Fu and Peixian Chen and Yunhang Shen and Yulei Qin and Mengdan Zhang and Xu Lin and Jinrui Yang and Xiawu Zheng and Ke Li and Xing Sun and Yunsheng Wu and Rongrong Ji and Caifeng Shan and Ran He},
      year={2025},
      eprint={2306.13394},
      archivePrefix={arXiv},
      primaryClass={cs.CV},
      url={https://arxiv.org/abs/2306.13394}, 
}

@inproceedings{hudson2019gqa,
  title={Gqa: A new dataset for real-world visual reasoning and compositional question answering},
  author={Hudson, Drew A and Manning, Christopher D},
  booktitle={Proceedings of the IEEE/CVF conference on computer vision and pattern recognition},
  pages={6700--6709},
  year={2019}
}

@online{xai2024-grok15v,
  author  = {{xAI}},
  title   = {Grok-1.5 Vision Preview},
  date    = {2024-04-12},
  url     = {https://x.ai/news/grok-1.5v},
  urldate = {2025-11-10},
  note    = {Company news post}
}

@inproceedings{yue2024mmmu,
  title={Mmmu: A massive multi-discipline multimodal understanding and reasoning benchmark for expert agi},
  author={Yue, Xiang and Ni, Yuansheng and Zhang, Kai and Zheng, Tianyu and Liu, Ruoqi and Zhang, Ge and Stevens, Samuel and Jiang, Dongfu and Ren, Weiming and Sun, Yuxuan and others},
  booktitle={Proceedings of the IEEE/CVF Conference on Computer Vision and Pattern Recognition},
  pages={9556--9567},
  year={2024}
}

@article{chen2024we,
  title={Are we on the right way for evaluating large vision-language models?},
  author={Chen, Lin and Li, Jinsong and Dong, Xiaoyi and Zhang, Pan and Zang, Yuhang and Chen, Zehui and Duan, Haodong and Wang, Jiaqi and Qiao, Yu and Lin, Dahua and others},
  journal={Advances in Neural Information Processing Systems},
  volume={37},
  pages={27056--27087},
  year={2024}
}

@inproceedings{kembhavi2016diagram,
  title={A diagram is worth a dozen images},
  author={Kembhavi, Aniruddha and Salvato, Mike and Kolve, Eric and Seo, Minjoon and Hajishirzi, Hannaneh and Farhadi, Ali},
  booktitle={European conference on computer vision},
  pages={235--251},
  year={2016},
  organization={Springer}
}

@inproceedings{masry2022chartqa,
  title={Chartqa: A benchmark for question answering about charts with visual and logical reasoning},
  author={Masry, Ahmed and Do, Xuan Long and Tan, Jia Qing and Joty, Shafiq and Hoque, Enamul},
  booktitle={Findings of the association for computational linguistics: ACL 2022},
  pages={2263--2279},
  year={2022}
}

@article{liu2024ocrbench,
  title={Ocrbench: on the hidden mystery of ocr in large multimodal models},
  author={Liu, Yuliang and Li, Zhang and Huang, Mingxin and Yang, Biao and Yu, Wenwen and Li, Chunyuan and Yin, Xu-Cheng and Liu, Cheng-Lin and Jin, Lianwen and Bai, Xiang},
  journal={Science China Information Sciences},
  volume={67},
  number={12},
  pages={220102},
  year={2024},
  publisher={Springer}
}

@article{seawead2025seaweed,
  title={Seaweed-7b: Cost-effective training of video generation foundation model},
  author={Seawead, Team and Yang, Ceyuan and Lin, Zhijie and Zhao, Yang and Lin, Shanchuan and Ma, Zhibei and Guo, Haoyuan and Chen, Hao and Qi, Lu and Wang, Sen and others},
  journal={arXiv preprint arXiv:2504.08685},
  year={2025}
}

@article{ghosh2023geneval,
  title={Geneval: An object-focused framework for evaluating text-to-image alignment},
  author={Ghosh, Dhruba and Hajishirzi, Hannaneh and Schmidt, Ludwig},
  journal={Advances in Neural Information Processing Systems},
  volume={36},
  pages={52132--52152},
  year={2023}
}

@article{loshchilov2017decoupled,
  title={Decoupled weight decay regularization},
  author={Loshchilov, Ilya and Hutter, Frank},
  journal={arXiv preprint arXiv:1711.05101},
  year={2017}
}

@article{wu2025qwen,
  title={Qwen-image technical report},
  author={Wu, Chenfei and Li, Jiahao and Zhou, Jingren and Lin, Junyang and Gao, Kaiyuan and Yan, Kun and Yin, Sheng-ming and Bai, Shuai and Xu, Xiao and Chen, Yilei and others},
  journal={arXiv preprint arXiv:2508.02324},
  year={2025}
}

@article{wu2025omnigen2,
  title={OmniGen2: Exploration to Advanced Multimodal Generation},
  author={Wu, Chenyuan and Zheng, Pengfei and Yan, Ruiran and Xiao, Shitao and Luo, Xin and Wang, Yueze and Li, Wanli and Jiang, Xiyan and Liu, Yexin and Zhou, Junjie and others},
  journal={arXiv preprint arXiv:2506.18871},
  year={2025}
}

@inproceedings{xiao2025omnigen,
  title={Omnigen: Unified image generation},
  author={Xiao, Shitao and Wang, Yueze and Zhou, Junjie and Yuan, Huaying and Xing, Xingrun and Yan, Ruiran and Li, Chaofan and Wang, Shuting and Huang, Tiejun and Liu, Zheng},
  booktitle={Proceedings of the Computer Vision and Pattern Recognition Conference},
  pages={13294--13304},
  year={2025}
}

@article{ye2025imgedit,
  title={Imgedit: A unified image editing dataset and benchmark},
  author={Ye, Yang and He, Xianyi and Li, Zongjian and Lin, Bin and Yuan, Shenghai and Yan, Zhiyuan and Hou, Bohan and Yuan, Li},
  journal={arXiv preprint arXiv:2505.20275},
  year={2025}
}

@article{hu2024ella,
  title={Ella: Equip diffusion models with llm for enhanced semantic alignment},
  author={Hu, Xiwei and Wang, Rui and Fang, Yixiao and Fu, Bin and Cheng, Pei and Yu, Gang},
  journal={arXiv preprint arXiv:2403.05135},
  year={2024}
}

@inproceedings{li2024mvbench,
  title={Mvbench: A comprehensive multi-modal video understanding benchmark},
  author={Li, Kunchang and Wang, Yali and He, Yinan and Li, Yizhuo and Wang, Yi and Liu, Yi and Wang, Zun and Xu, Jilan and Chen, Guo and Luo, Ping and others},
  booktitle={Proceedings of the IEEE/CVF Conference on Computer Vision and Pattern Recognition},
  pages={22195--22206},
  year={2024}
}

@article{wu2024longvideobench,
  title={Longvideobench: A benchmark for long-context interleaved video-language understanding},
  author={Wu, Haoning and Li, Dongxu and Chen, Bei and Li, Junnan},
  journal={Advances in Neural Information Processing Systems},
  volume={37},
  pages={28828--28857},
  year={2024}
}

@inproceedings{huang2024vbench,
  title={Vbench: Comprehensive benchmark suite for video generative models},
  author={Huang, Ziqi and He, Yinan and Yu, Jiashuo and Zhang, Fan and Si, Chenyang and Jiang, Yuming and Zhang, Yuanhan and Wu, Tianxing and Jin, Qingyang and Chanpaisit, Nattapol and others},
  booktitle={Proceedings of the IEEE/CVF Conference on Computer Vision and Pattern Recognition},
  pages={21807--21818},
  year={2024}
}

@article{han2025vision,
  title={Vision as a Dialect: Unifying Visual Understanding and Generation via Text-Aligned Representations},
  author={Han, Jiaming and Chen, Hao and Zhao, Yang and Wang, Hanyu and Zhao, Qi and Yang, Ziyan and He, Hao and Yue, Xiangyu and Jiang, Lu},
  journal={arXiv preprint arXiv:2506.18898},
  year={2025}
}

@article{geng2025x,
  title={X-omni: Reinforcement learning makes discrete autoregressive image generative models great again},
  author={Geng, Zigang and Wang, Yibing and Ma, Yeyao and Li, Chen and Rao, Yongming and Gu, Shuyang and Zhong, Zhao and Lu, Qinglin and Hu, Han and Zhang, Xiaosong and others},
  journal={arXiv preprint arXiv:2507.22058},
  year={2025}
}

@article{su2024roformer,
  title={Roformer: Enhanced transformer with rotary position embedding},
  author={Su, Jianlin and Ahmed, Murtadha and Lu, Yu and Pan, Shengfeng and Bo, Wen and Liu, Yunfeng},
  journal={Neurocomputing},
  volume={568},
  pages={127063},
  year={2024},
  publisher={Elsevier}
}

@article{li2023seed,
  title={Seed-bench: Benchmarking multimodal llms with generative comprehension},
  author={Li, Bohao and Wang, Rui and Wang, Guangzhi and Ge, Yuying and Ge, Yixiao and Shan, Ying},
  journal={arXiv preprint arXiv:2307.16125},
  year={2023}
}

@article{chang2025oneig,
  title={OneIG-Bench: Omni-dimensional Nuanced Evaluation for Image Generation},
  author={Chang, Jingjing and Fang, Yixiao and Xing, Peng and Wu, Shuhan and Cheng, Wei and Wang, Rui and Zeng, Xianfang and Yu, Gang and Chen, Hai-Bao},
  journal={arXiv preprint arXiv:2506.07977},
  year={2025}
}

@article{alayrac2022flamingo,
  title={Flamingo: a visual language model for few-shot learning},
  author={Alayrac, Jean-Baptiste and Donahue, Jeff and Luc, Pauline and Miech, Antoine and Barr, Iain and Hasson, Yana and Lenc, Karel and Mensch, Arthur and Millican, Katherine and Reynolds, Malcolm and others},
  journal={Advances in neural information processing systems},
  volume={35},
  pages={23716--23736},
  year={2022}
}

@article{laurenccon2023obelics,
  title={Obelics: An open web-scale filtered dataset of interleaved image-text documents},
  author={Lauren{\c{c}}on, Hugo and Saulnier, Lucile and Tronchon, L{\'e}o and Bekman, Stas and Singh, Amanpreet and Lozhkov, Anton and Wang, Thomas and Karamcheti, Siddharth and Rush, Alexander and Kiela, Douwe and others},
  journal={Advances in Neural Information Processing Systems},
  volume={36},
  pages={71683--71702},
  year={2023}
}

@article{an2025llava,
  title={Llava-onevision-1.5: Fully open framework for democratized multimodal training},
  author={An, Xiang and Xie, Yin and Yang, Kaicheng and Zhang, Wenkang and Zhao, Xiuwei and Cheng, Zheng and Wang, Yirui and Xu, Songcen and Chen, Changrui and Wu, Chunsheng and others},
  journal={arXiv preprint arXiv:2509.23661},
  year={2025}
}

@misc{llavanext2024,
  title        = {LLAVA-Next},
  author = {Liu, Haotian and Li, Chunyuan and Li, Yuheng and Li, Bo and Zhang, Yuanhan and Shen, Sheng and Lee, Yong Jae},
  year         = {2024},
  howpublished = {\url{https://llava-vl.github.io/blog/2024-01-30-llava-next/}},
  note         = {Accessed: 2025-02-14}
}

@inproceedings{chen2024sharegpt4v,
  title={Sharegpt4v: Improving large multi-modal models with better captions},
  author={Chen, Lin and Li, Jinsong and Dong, Xiaoyi and Zhang, Pan and He, Conghui and Wang, Jiaqi and Zhao, Feng and Lin, Dahua},
  booktitle={European Conference on Computer Vision},
  pages={370--387},
  year={2024},
  organization={Springer}
}

@article{ren2024vista,
  title={VISTA: Enhancing Long-Duration and High-Resolution Video Understanding by Video Spatiotemporal Augmentation},
  author={Ren, Weiming and Yang, Huan and Min, Jie and Wei, Cong and Chen, Wenhu},
  journal={arXiv preprint arXiv:2412.00927},
  year={2024}
}

@article{laurenccon2024matters,
  title={What matters when building vision-language models?},
  author={Lauren{\c{c}}on, Hugo and Tronchon, L{\'e}o and Cord, Matthieu and Sanh, Victor},
  journal={arXiv preprint arXiv:2405.02246},
  year={2024}
}

@article{jiang2024mantis,
  title={Mantis: Interleaved multi-image instruction tuning},
  author={Jiang, Dongfu and He, Xuan and Zeng, Huaye and Wei, Cong and Ku, Max and Liu, Qian and Chen, Wenhu},
  journal={arXiv preprint arXiv:2405.01483},
  year={2024}
}

@article{lin2023video,
  title={Video-llava: Learning united visual representation by alignment before projection},
  author={Lin, Bin and Ye, Yang and Zhu, Bin and Cui, Jiaxi and Ning, Munan and Jin, Peng and Yuan, Li},
  journal={arXiv preprint arXiv:2311.10122},
  year={2023}
}

@article{wang2024qwen2,
  title={Qwen2-vl: Enhancing vision-language model's perception of the world at any resolution},
  author={Wang, Peng and Bai, Shuai and Tan, Sinan and Wang, Shijie and Fan, Zhihao and Bai, Jinze and Chen, Keqin and Liu, Xuejing and Wang, Jialin and Ge, Wenbin and others},
  journal={arXiv preprint arXiv:2409.12191},
  year={2024}
}

@article{zhang2024long,
  title={Long context transfer from language to vision},
  author={Zhang, Peiyuan and Zhang, Kaichen and Li, Bo and Zeng, Guangtao and Yang, Jingkang and Zhang, Yuanhan and Wang, Ziyue and Tan, Haoran and Li, Chunyuan and Liu, Ziwei},
  journal={arXiv preprint arXiv:2406.16852},
  year={2024}
}

@article{maaz2023video,
  title={Video-chatgpt: Towards detailed video understanding via large vision and language models},
  author={Maaz, Muhammad and Rasheed, Hanoona and Khan, Salman and Khan, Fahad Shahbaz},
  journal={arXiv preprint arXiv:2306.05424},
  year={2023}
}

@inproceedings{deng2025openvlthinker,
  title={Openvlthinker: Complex vision-language reasoning via iterative sft-rl cycles},
  author={Deng, Yihe and Bansal, Hritik and Yin, Fan and Peng, Nanyun and Wang, Wei and Chang, Kai-Wei},
  booktitle={The Thirty-ninth Annual Conference on Neural Information Processing Systems},
  year={2025}
}

@article{huang2025vision,
  title={Vision-r1: Incentivizing reasoning capability in multimodal large language models},
  author={Huang, Wenxuan and Jia, Bohan and Zhai, Zijie and Cao, Shaosheng and Ye, Zheyu and Zhao, Fei and Xu, Zhe and Hu, Yao and Lin, Shaohui},
  journal={arXiv preprint arXiv:2503.06749},
  year={2025}
}

@article{feng2025video,
  title={Video-r1: Reinforcing video reasoning in mllms},
  author={Feng, Kaituo and Gong, Kaixiong and Li, Bohao and Guo, Zonghao and Wang, Yibing and Peng, Tianshuo and Wu, Junfei and Zhang, Xiaoying and Wang, Benyou and Yue, Xiangyu},
  journal={arXiv preprint arXiv:2503.21776},
  year={2025}
}

@article{su2025pixel,
  title={Pixel reasoner: Incentivizing pixel-space reasoning with curiosity-driven reinforcement learning},
  author={Su, Alex and Wang, Haozhe and Ren, Weiming and Lin, Fangzhen and Chen, Wenhu},
  journal={arXiv preprint arXiv:2505.15966},
  year={2025}
}

@article{su2025thinking,
  title={Thinking with images for multimodal reasoning: Foundations, methods, and future frontiers},
  author={Su, Zhaochen and Xia, Peng and Guo, Hangyu and Liu, Zhenhua and Ma, Yan and Qu, Xiaoye and Liu, Jiaqi and Li, Yanshu and Zeng, Kaide and Yang, Zhengyuan and others},
  journal={arXiv preprint arXiv:2506.23918},
  year={2025}
}

@article{liu2025visionreasoner,
  title={VisionReasoner: Unified Visual Perception and Reasoning via Reinforcement Learning},
  author={Liu, Yuqi and Qu, Tianyuan and Zhong, Zhisheng and Peng, Bohao and Liu, Shu and Yu, Bei and Jia, Jiaya},
  journal={arXiv preprint arXiv:2505.12081},
  year={2025}
}

@article{van2017neural,
  title={Neural discrete representation learning},
  author={Van Den Oord, Aaron and Vinyals, Oriol and others},
  journal={Advances in neural information processing systems},
  volume={30},
  year={2017}
}

@article{dhariwal2021diffusion,
  title={Diffusion models beat gans on image synthesis},
  author={Dhariwal, Prafulla and Nichol, Alexander},
  journal={Advances in neural information processing systems},
  volume={34},
  pages={8780--8794},
  year={2021}
}

@article{podell2023sdxl,
  title={Sdxl: Improving latent diffusion models for high-resolution image synthesis},
  author={Podell, Dustin and English, Zion and Lacey, Kyle and Blattmann, Andreas and Dockhorn, Tim and M{\"u}ller, Jonas and Penna, Joe and Rombach, Robin},
  journal={arXiv preprint arXiv:2307.01952},
  year={2023}
}

@article{saharia2022photorealistic,
  title={Photorealistic text-to-image diffusion models with deep language understanding},
  author={Saharia, Chitwan and Chan, William and Saxena, Saurabh and Li, Lala and Whang, Jay and Denton, Emily L and Ghasemipour, Kamyar and Gontijo Lopes, Raphael and Karagol Ayan, Burcu and Salimans, Tim and others},
  journal={Advances in neural information processing systems},
  volume={35},
  pages={36479--36494},
  year={2022}
}

@inproceedings{ronneberger2015u,
  title={U-net: Convolutional networks for biomedical image segmentation},
  author={Ronneberger, Olaf and Fischer, Philipp and Brox, Thomas},
  booktitle={International Conference on Medical image computing and computer-assisted intervention},
  pages={234--241},
  year={2015},
  organization={Springer}
}

@article{ho2020denoising,
  title={Denoising diffusion probabilistic models},
  author={Ho, Jonathan and Jain, Ajay and Abbeel, Pieter},
  journal={Advances in neural information processing systems},
  volume={33},
  pages={6840--6851},
  year={2020}
}

@inproceedings{ma2024sit,
  title={Sit: Exploring flow and diffusion-based generative models with scalable interpolant transformers},
  author={Ma, Nanye and Goldstein, Mark and Albergo, Michael S and Boffi, Nicholas M and Vanden-Eijnden, Eric and Xie, Saining},
  booktitle={European Conference on Computer Vision},
  pages={23--40},
  year={2024},
  organization={Springer}
}

@article{song2020score,
  title={Score-based generative modeling through stochastic differential equations},
  author={Song, Yang and Sohl-Dickstein, Jascha and Kingma, Diederik P and Kumar, Abhishek and Ermon, Stefano and Poole, Ben},
  journal={arXiv preprint arXiv:2011.13456},
  year={2020}
}

@article{liu2022flow,
  title={Flow straight and fast: Learning to generate and transfer data with rectified flow},
  author={Liu, Xingchao and Gong, Chengyue and Liu, Qiang},
  journal={arXiv preprint arXiv:2209.03003},
  year={2022}
}

@article{lipman2022flow,
  title={Flow matching for generative modeling},
  author={Lipman, Yaron and Chen, Ricky TQ and Ben-Hamu, Heli and Nickel, Maximilian and Le, Matt},
  journal={arXiv preprint arXiv:2210.02747},
  year={2022}
}

@article{albergo2023stochastic,
  title={Stochastic interpolants: A unifying framework for flows and diffusions},
  author={Albergo, Michael S and Boffi, Nicholas M and Vanden-Eijnden, Eric},
  journal={arXiv preprint arXiv:2303.08797},
  year={2023}
}

@article{li2024hunyuan,
  title={Hunyuan-dit: A powerful multi-resolution diffusion transformer with fine-grained chinese understanding},
  author={Li, Zhimin and Zhang, Jianwei and Lin, Qin and Xiong, Jiangfeng and Long, Yanxin and Deng, Xinchi and Zhang, Yingfang and Liu, Xingchao and Huang, Minbin and Xiao, Zedong and others},
  journal={arXiv preprint arXiv:2405.08748},
  year={2024}
}

@article{kong2024hunyuanvideo,
  title={Hunyuanvideo: A systematic framework for large video generative models},
  author={Kong, Weijie and Tian, Qi and Zhang, Zijian and Min, Rox and Dai, Zuozhuo and Zhou, Jin and Xiong, Jiangfeng and Li, Xin and Wu, Bo and Zhang, Jianwei and others},
  journal={arXiv preprint arXiv:2412.03603},
  year={2024}
}

@inproceedings{fu2025video,
  title={Video-mme: The first-ever comprehensive evaluation benchmark of multi-modal llms in video analysis},
  author={Fu, Chaoyou and Dai, Yuhan and Luo, Yongdong and Li, Lei and Ren, Shuhuai and Zhang, Renrui and Wang, Zihan and Zhou, Chenyu and Shen, Yunhang and Zhang, Mengdan and others},
  booktitle={Proceedings of the Computer Vision and Pattern Recognition Conference},
  pages={24108--24118},
  year={2025}
}

@inproceedings{wang2025lvbench,
  title={Lvbench: An extreme long video understanding benchmark},
  author={Wang, Weihan and He, Zehai and Hong, Wenyi and Cheng, Yean and Zhang, Xiaohan and Qi, Ji and Ding, Ming and Gu, Xiaotao and Huang, Shiyu and Xu, Bin and others},
  booktitle={Proceedings of the IEEE/CVF International Conference on Computer Vision},
  pages={22958--22967},
  year={2025}
}

@misc{OpenAI_GPT4o,
  author       = {OpenAI},
  title        = {GPT-4o},
  year         = {2024},
  howpublished = {\url{https://openai.com/index/hello-gpt-4o/}},
}

@article{team2024gemini,
  title={Gemini 1.5: Unlocking multimodal understanding across millions of tokens of context},
  author={Team, Gemini and Georgiev, Petko and Lei, Ving Ian and Burnell, Ryan and Bai, Libin and Gulati, Anmol and Tanzer, Garrett and Vincent, Damien and Pan, Zhufeng and Wang, Shibo and others},
  journal={arXiv preprint arXiv:2403.05530},
  year={2024}
}

@article{cheng2024videollama,
  title={Videollama 2: Advancing spatial-temporal modeling and audio understanding in video-llms},
  author={Cheng, Zesen and Leng, Sicong and Zhang, Hang and Xin, Yifei and Li, Xin and Chen, Guanzheng and Zhu, Yongxin and Zhang, Wenqi and Luo, Ziyang and Zhao, Deli and others},
  journal={arXiv preprint arXiv:2406.07476},
  year={2024}
}

@article{luo2025next,
  title={NExT-OMNI: Towards Any-to-Any Omnimodal Foundation Models with Discrete Flow Matching},
  author={Luo, Run and Xia, Xiaobo and Wang, Lu and Chen, Longze and Shan, Renke and Luo, Jing and Yang, Min and Chua, Tat-Seng},
  journal={arXiv preprint arXiv:2510.13721},
  year={2025}
}

@article{liu2025worldweaver,
 title={Worldweaver: Generating long-horizon video worlds via rich perception},
 author={Liu, Zhiheng and Deng, Xueqing and Chen, Shoufa and Wang, Angtian and Guo, Qiushan and Han, Mingfei and Xue, Zeyue and Chen, Mengzhao and Luo, Ping and Yang, Linjie},
 journal={arXiv preprint arXiv:2508.15720},
 year={2025}
}

@inproceedings{liu2025manganinja,
 title={Manganinja: Line art colorization with precise reference following},
 author={Liu, Zhiheng and Cheng, Ka Leong and Chen, Xi and Xiao, Jie and Ouyang, Hao and Zhu, Kai and Liu, Yu and Shen, Yujun and Chen, Qifeng and Luo, Ping},
 booktitle={Proceedings of the Computer Vision and Pattern Recognition Conference},
 pages={5666--5677},
 year={2025}
}

@article{liu2023cones,
 title={Cones: Concept neurons in diffusion models for customized generation},
 author={Liu, Zhiheng and Feng, Ruili and Zhu, Kai and Zhang, Yifei and Zheng, Kecheng and Liu, Yu and Zhao, Deli and Zhou, Jingren and Cao, Yang},
 journal={arXiv preprint arXiv:2303.05125},
 year={2023}
}

@article{liu2023customizable,
 title={Customizable image synthesis with multiple subjects},
 author={Liu, Zhiheng and Zhang, Yifei and Shen, Yujun and Zheng, Kecheng and Zhu, Kai and Feng, Ruili and Liu, Yu and Zhao, Deli and Zhou, Jingren and Cao, Yang},
 journal={Advances in neural information processing systems},
 volume={36},
 pages={57500--57519},
 year={2023}
}

@article{ren2025vamba,
  title={Vamba: Understanding hour-long videos with hybrid mamba-transformers},
  author={Ren, Weiming and Ma, Wentao and Yang, Huan and Wei, Cong and Zhang, Ge and Chen, Wenhu},
  journal={arXiv preprint arXiv:2503.11579},
  year={2025}
}

\end{document}